\documentclass[final]{cvpr}

\usepackage{graphicx}
\usepackage{comment}
\usepackage{amsmath,amssymb} 
\usepackage{color}
\usepackage{bm}
\usepackage{diagbox}

\usepackage{verbatim}
\usepackage{multirow}
\usepackage{makecell}
\usepackage{amsmath}
\usepackage{subfigure}

\usepackage{times}
\usepackage{epsfig}
\usepackage{graphicx}
\usepackage{amsmath}
\usepackage{amssymb}

\DeclareMathOperator*{\Exp}{\mathbb{E}}

\usepackage[pagebackref=true,breaklinks=true,colorlinks,bookmarks=false]{hyperref}

\begin{document}

\title{BCNet: Searching for Network Width with Bilaterally Coupled Network}

\author{Xiu Su$^{1}$, Shan You$^{2,3}$\thanks{Corresponding authors.}, ~Fei Wang$^{2}$, Chen Qian$^{2}$, Changshui Zhang$^{3}$, Chang Xu$^{1*}$\\
$^1$School of Computer Science, Faculty of Engineering, The University of Sydney, Australia\\
$^2$SenseTime Research\\
$^3$Department of Automation, Tsinghua University,\\
Institute for Artificial Intelligence, Tsinghua University (THUAI), \\
Beijing National Research Center for Information Science and Technology (BNRist) \\
{\tt\small xisu5992@uni.sydney.edu.au, \{youshan,wangfei,qianchen\}@sensetime.com}\\
{\tt\small zcs@mail.tsinghua.edu.cn, c.xu@sydney.edu.au}
}

\maketitle

\newcommand{\fracpartial}[2]{\frac{\partial #1}{\partial  #2}}
\newcommand{\norm}[1]{\left\lVert#1\right\rVert}
\newcommand{\innerproduct}[2]{\left\langle#1, #2\right\rangle}
\newcommand{\fan}[1]{\Vert #1 \Vert}
\newcommand{\qileft}{[\kern-0.15em[}
\newcommand{\qiLeft}{\left[\kern-0.4em\left[}
\newcommand{\qiright}{]\kern-0.15em]}
\newcommand{\qiRight}{\right]\kern-0.4em\right]}
\newcommand{\sign}{{\mbox{sign}}}
\newcommand{\diag}{{\mbox{diag}}}
\newcommand{\armin}{{\mbox{argmin}}}
\newcommand{\rank}{{\mbox{rank}}}
\renewcommand{\vec}{{\mbox{vec}}}
\newcommand{\st}{{\mbox{s.t.}}}
\newcommand{\<}{\left\langle}
\renewcommand{\>}{\right\rangle}
\newcommand{\lbar}{\left\|}
\newcommand{\rbar}{\right\|}
\renewcommand{\Roman}[1]{\uppercase\expandafter{\romannumeral#1}}
\newcommand{\red}[1]{{\color{red}{#1}}}
\newcommand{\blue}[1]{{\color{blue}{#1}}}
\newcommand{\FLOPs}{\mbox{FLOPs}}
\newcommand{\FC}{\mbox{FC}}
\newcommand{\Sigmoid}{\mbox{Sigmoid}}
\newcommand{\ReLU}{\mbox{ReLU}}

\renewcommand{\a}{{\bm{a}}}
\renewcommand{\b}{{\bm{b}}}
\renewcommand{\c}{{\bm{c}}}
\renewcommand{\d}{{\bm{d}}}
\newcommand{\e}{{\bm{e}}}
\newcommand{\f}{{\bm{f}}}
\newcommand{\g}{{\bm{g}}}
\renewcommand{\o}{{\bm{o}}}
\newcommand{\p}{{\bm{p}}}
\newcommand{\q}{{\bm{q}}}
\newcommand{\s}{{\bm{s}}}
\renewcommand{\t}{{\bm{t}}}
\renewcommand{\u}{{\bm{u}}}
\renewcommand{\v}{{\bm{v}}}
\newcommand{\w}{{\bm{w}}}
\newcommand{\x}{{\bm{x}}}
\newcommand{\y}{{\bm{y}}}
\newcommand{\z}{{\bm{z}}}
\newcommand{\balpha}{{\bm{\alpha}}}
\newcommand{\bbeta}{{\bm{\beta}}}
\newcommand{\bmu}{{\bm{\mu}}}
\newcommand{\bsigma}{{\bm{\sigma}}}
\newcommand{\blambda}{{\bm{\lambda}}}
\newcommand{\bgamma}{{\bm{\gamma}}}
\newcommand{\bxi}{{\bm{\xi}}}
\newcommand{\bphi}{{\bm{\phi}}}

\newcommand{\ba}{{\bm{A}}}
\newcommand{\bb}{{\bm{B}}}
\newcommand{\bc}{{\bm{C}}}
\newcommand{\bd}{{\bm{D}}}
\newcommand{\be}{{\bm{E}}}
\newcommand{\bg}{{\bm{G}}}
\newcommand{\bi}{{\bm{I}}}
\newcommand{\bj}{{\bm{J}}}
\newcommand{\bl}{{\bm{L}}}
\newcommand{\bo}{{\bm{O}}}
\newcommand{\bp}{{\bm{P}}}
\newcommand{\bq}{{\bm{Q}}}
\newcommand{\bs}{{\bm{S}}}
\newcommand{\bu}{{\bm{U}}}
\newcommand{\bv}{{\bm{V}}}
\newcommand{\bw}{{\bm{W}}}
\newcommand{\bx}{{\bm{X}}}
\newcommand{\by}{{\bm{Y}}}
\newcommand{\bz}{{\bm{Z}}}
\newcommand{\bTheta}{{\bm{\Theta}}}
\newcommand{\bSigma}{{\bm{\Sigma}}}

\newcommand{\A}{{\mathcal{A}}}
\newcommand{\B}{\mathcal{B}}
\newcommand{\C}{\mathcal{C}}
\newcommand{\D}{\mathcal{D}}
\newcommand{\E}{\mathcal{E}}
\newcommand{\F}{\mathcal{F}}
\renewcommand{\H}{\mathcal{H}}
\newcommand{\I}{\mathcal{I}}
\renewcommand{\L}{\mathcal{L}}
\newcommand{\N}{\mathcal{N}}
\renewcommand{\P}{\mathcal{P}}
\newcommand{\X}{\mathcal{X}}
\newcommand{\Y}{\mathcal{Y}}
\newcommand{\W}{\mathcal{W}}

\begin{abstract}
Searching for a more compact network width recently serves as an effective way of channel pruning for the deployment of convolutional neural networks (CNNs) under hardware constraints. To fulfill the searching, a one-shot supernet is usually leveraged to efficiently evaluate the performance \wrt~different network widths. However, current methods mainly follow a \textit{unilaterally augmented} (UA) principle for the evaluation of each width, which induces the training unfairness of channels in supernet. In this paper, we introduce a new supernet called Bilaterally Coupled Network (BCNet) to address this issue. In BCNet, each channel is fairly trained and responsible for the same amount of network widths, thus each network width can be evaluated more accurately. Besides, we leverage a stochastic complementary strategy for training the BCNet, and propose a prior initial population sampling method to boost the performance of the evolutionary search. Extensive experiments on benchmark CIFAR-10 and ImageNet datasets indicate that our method can achieve state-of-the-art or competing performance over other baseline methods. Moreover, our method turns out to further boost the performance of NAS models by refining their network widths. For example, with the same FLOPs budget, our obtained EfficientNet-B0 achieves 77.36\% Top-1 accuracy on ImageNet dataset, surpassing the performance of original setting by 0.48\%.

\end{abstract}

\section{Introduction}

For practical deployment of convolutional neural networks (CNNs), it is important to consider different hardware budgets \cite{dc,han2020ghostnet,han2020model,mobilenetv2}, to name a few, floating point operations (FLOPs), latency, memory footprint and energy consumption. One way to simultaneously accommodate all these budgets is to prune the redundant channels of a model, so that a compact network width can be obtained. Typical channel pruning usually leverages a pre-trained network and implement the pruning in an end-to-end \cite{sss,slimming,su2020data} or layer-by-layer \cite{cp,tang2020reborn} manner. After pruning, the structure of the pre-trained model remains unchanged, so that the pruned network is friendly to off-the-shelf deep learning frameworks and can be further boosted by other techniques, such as quantization \cite{dc} and knowledge distillation \cite{distilling,you2017learning,kong2020learning}.

Recently, \cite{rethinkingpruning} found the core of channel pruning is to learn a more compact \textit{network width} instead of the retained weights. Other literature also uses number of channels/filters to indicate the network width. Thus follow-up work resorts to neural architecture search (NAS) \cite{yang2020ista,you2020greedynas,yang2021towards,proxylessnas} or other automl techniques for directly searching for an optimal network width, such as 
MetaPruning \cite{metapruning}, AutoSlim \cite{autoslim} and TAS \cite{tas}. In their methods, a one-shot supernet is usually leveraged for evaluation of different widths. Concretely, for the width $c$ at a certain layer, we need to assign $c$ channels (filters) in the layer and all layers follow the same way. Then all these assigned channels in the supernet specify a sub-network with the supernet. As a result, the performance of a network width refers to the accuracy of the specified sub-network with shared weights of supernet. For fair evaluation of different network widths, during the training of supernet, all network widths will be evenly sampled from the supernet and get optimized accordingly.  For brevity, we use the name of layer width to indicate the width for a certain layer, while network width represents the set of widths for all layers.

\begin{figure*}[t]
	\centering
	\includegraphics[width=0.90\linewidth]{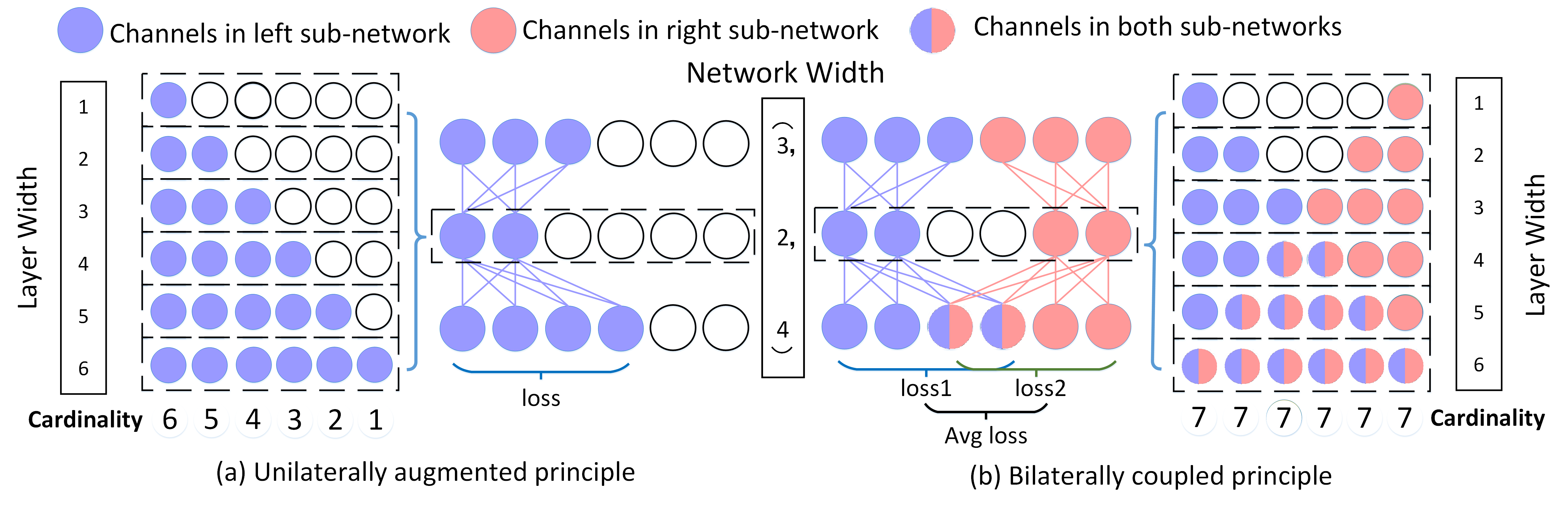}
	\vspace{-2mm}
	\caption{Comparison of unilaterally augmented (UA) principle and our proposed bilaterally coupled (BC) principle in supernet. In BC principle, each network width is indicated by two (left and right) paths, so that all channels get the same cardinality for evaluation different widths. However, in UA principle each width goes through one path, and training unfairness over channels and evaluation bias exist. Under uniform sampling strategy, for each channel the expectation of the times being evaluated is theoretically equal to the times being trained,  since we simply sample each path and train it. For simplicity,  we use \emph{cardinality} to refer to the number of times that a channel is  used for evaluation over all widths.}
	\label{motivation}
	\vspace{-5mm}
\end{figure*}

In this way, how to specify the sub-network(s) for each network width matters for the performance evaluation. However, current methods \cite{metapruning,autoslim,tas} mainly follow a \textit{unilaterally augmented} (UA) principle for the evaluation of network widths in supernet. Suppose we count channels in a layer from the left to the right as Figure \ref{motivation}. To evaluate the width $c$, UA principle simply assigns the leftmost $c$ channels to specify a sub-network for evaluation. In this way, channels within smaller width will also be used for evaluation of larger widths. Since we uniformly sample all widths during training the supernet, channels close to left side will be used more times than those close to the right side in the evaluation of widths as in Figure \ref{motivation}(a). For example, the leftmost channel will be used 6 times for evaluation while the rightmost channel is only used once. This causes \textit{training unfairness} among the channels and their corresponding kernels. Left channels will be trained more than right ones. Nevertheless, this training unfairness will affect the accuracy of evaluation, and thus hampers the ability of supernet to rank over all network widths.

In this paper, we introduce a new supernet called Bilaterally Coupled Network (BCNet) to address the training and evaluation unfairness within UA principle. In BCNet, each channel is fairly trained and responsible for the same amount of widths. Specifically, both in training and evaluation, each width is determined symmetrically by the average performance of bilateral (\ie, both left and right) channels. As shown in Figure \ref{motivation}(b), suppose a layer has 6 channels, then each channel of BCNet evenly corresponds to 7 layer widths from the left or right side. In this way, all channels will be trained equally; all widths are bilaterally coupled in BCNet and will be evaluated more fairly.

To encourage a rigorous training fairness over channels, we adopt a complementary training strategy for training BCNet as in Figure \ref{complementary}. As for the subsequent searching, since the evolutionary algorithm is empirically fairly sensitive to the initial population, we also propose a prior sampling method, which enables to generate a good and steady initial population instead of random initialization. Extensive experiments on the benchmark CIFAR-10 and ImageNet datasets show that our method outperforms the state-of-the-art methods under various FLOPs budget. For example, our searched EfficientNet-B0 achieves 74.9\% Top-1 accuracy on ImageNet dataset with 192M FLOPs (2 $\times$ acceleration).

\section{Related Work}

Channel pruning is an effective method to compress and accelerate an over-parameterized convolutional neural network, and thus enables the pruned network to accommodate various hardware computational budgets. Extensive studies are illustrated in the comprehensive survey \cite{survey}. Here, we summarize the typical approaches of channel pruning \cite{sss,slimming,cp,tang2020scop} and network width search methods \cite{metapruning,autoslim,tas}.

\textbf{Channel pruning.} Channel pruning is an prevalent method which aims to reduce redundant channels of an heavy model, and generally implemented by selecting significant channels \cite{slimming,cp} or adding additional data-driven sparsity \cite{sss,dcp,tang2019bringing}. For example, CP \cite{cp} propose to construct a group Lasso to select unimportant channels. Slimming \cite{slimming} impose a $\l_1$ regularization on the scaling factors. DCP \cite{dcp} propose to construct an additional discrimination-aware losses. Despite the achievements, these methods rely heavily on manually assigned pruning ratios or hyperparameter coefficients, which is complicated, time consuming and hardly to find Oracle solutions.
 

\textbf{Network width search.} Inspired by the development of NAS \cite{proxylessnas,guo2020hit,huang2020explicitly}, network width search methods \cite{metapruning,autoslim,tas,dmcp,amc,su2021locally} generally take a carefully designed one-shot supernet to rank the relative  performance of different widths. For example, TAS \cite{tas} aims to search the optimal network width via a learnable continuous parameter distribution.  MetaPruning \cite{metapruning} proposes to directly generate representative weights for different widths. AutoSlim \cite{autoslim} proposes to leverage a slimmable network to approximate the accuracy of different network widths.  However, all these methods follow UA principle in assigning channels, which affects the fairness in evaluation. To accurately rank the performance of network widths, our proposed BCNet aims to assign the same opportunity for channels during training, thus ensures the evaluation fairness in searching optimal widths.

\section{Channel Pruning as Network Width Search}

Formally, suppose the target network to be pruned $\N$ has $L$ layers, and each layer has $l_i$ channels. Then channel pruning aims to identify redundant channels (indexed by $\I_{pruned}^i$) layer-wisely, \ie,
\begin{equation}
\I_{pruned}^i \subset [1:l_i],
\end{equation}
where $[1:l_i]$ is an index set for all integers in the range of $1$ to $l_i$ for $i$-th layer. However, \cite{rethinkingpruning} empirically finds that the absolute set of pruned channels $\I_{pruned}^i$ and their weights are not really necessary for the performance of pruned network, but the obtained width $c_i$ actually matters, \ie,
\begin{equation}
c_i = l_i - |\I_{pruned}^i|.
\end{equation}
In this way, it is intuitive to directly search for the optimal network width to meet the given budgets. 

Denote an arbritary network width as $\c = (c_1,c_2,...,c_L) \in \C = \bigotimes_{i=1}^L [1:l_i]$, where $\bigotimes$ is the Cartesian product. Then the size of search space $\C$ amounts to $|\C| = \prod_{i=1}^{L}l_i$. However, this search space is fairly huge, \eg,  $10^{25}$ for $L=25$ layers and $l_i=10$ channels. To reduce the search space, current methods tend to search on a group level instead of channel-wise level. In specific, all channels at a layer is partitioned evenly into $K$ groups, then we only need to consider $K$ cases; there are just $(l_i/K)\cdot[1:K]$ layer widths for $i$-th layer. Therefore, the search space $\C$ is shrunk into $\C_K$ with size $|\C_K| = K^L$. In the following, we use both $\C$ and $\C_K$ seamlessly. 


During searching, the target network is usually leveraged as a supernet $\N$, and different network widths $\c$ can be directly evaluated by sharing the same weights with the supernet. 
Then the width searching can be divided into two steps, i.e., supernet training, and searching with supernet. Usually, the original training dataset is split into two datasets, \ie, training dataset $\D_{tr}$ and validation dataset $\D_{val}$. The weights $\W$ of the target supernet $\N$ is trained by uniformly sampling a width $c$ and optimizing its corresponding sub-network with weights with weights $w_c \subset W$, 
\begin{equation}
W^* = \mathop{\arg\min}_{\w_\c \subset W}~ \Exp_{\c\in U(\C)} \qiLeft \L_{tr}(\w_c; \N, \c, \D_{tr})\qiRight, 
\label{eq2}
\end{equation}
where $\L_{train}$ is the training loss function, $U(\C)$ is a uniform distribution of network widths, 
and $\Exp\qiLeft\cdot\qiRight$ is the expectation
of random variables. Then the optimal network width $\c^*$ corresponds to the one with best performance on validation dataset, \eg~classification accuracy,
\begin{equation}
\begin{aligned}
\c^* = &\mathop{\arg\max}_{\c \in \C}~\mbox{Accuracy}(\c, \w^*_\c; W^*, \N, \D_{val}), \\ &~\st~\FLOPs(\c) \leq F_b,
\end{aligned}
\label{eq3}
\end{equation}	
where $F_b$ is a specified budget of FLOPs. Here we consider FLOPs rather than latency as the hardware
constraint since we are not targeting any specific hardware device like EfficientNet \cite{efficientnet} and other pruning baselines \cite{dcp,tas,amc}. The searching of Eq.\eqref{eq3} can be fulfilled efficiently by various algorithms, such as random or evolutionary search \cite{metapruning}. Afterwards, the performance of the searched optimal width $\c^*$ is analyzed by training from scratch.


\section{BCNet: Bilaterally Coupled Network}




\begin{figure*}[t]
	\centering
	\includegraphics[width=0.82\linewidth]{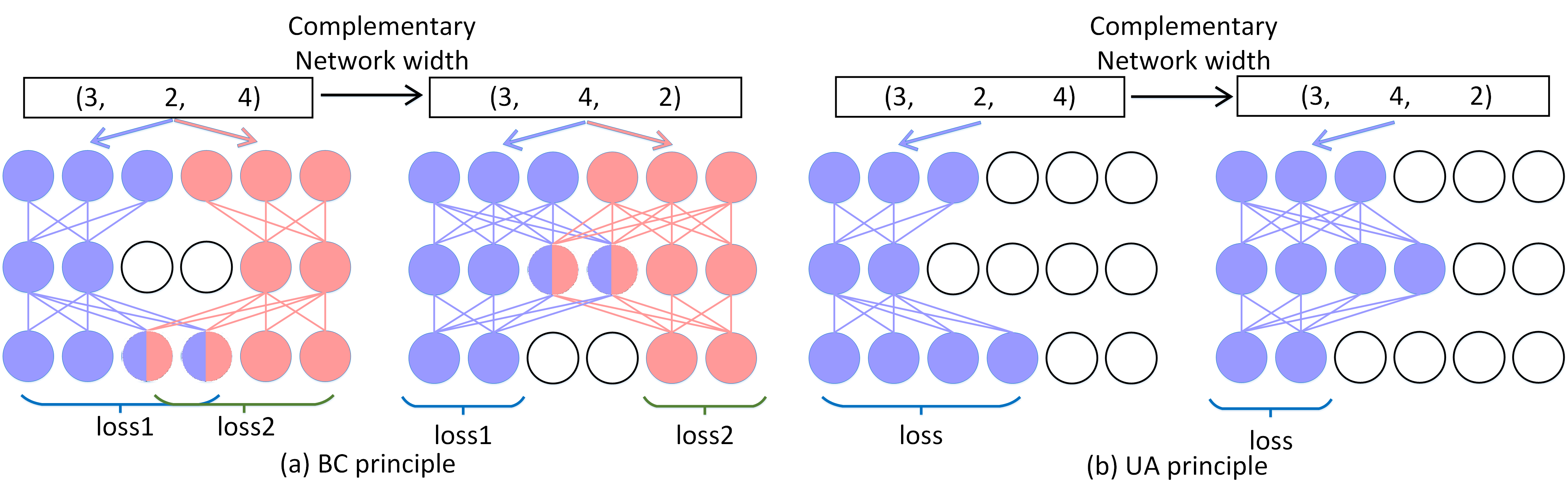}
	\caption{Illustration of the complementary of a network width for both our bilaterally coupled (BC) principle and the baseline unilaterally augmented (UA) principle. In BC principle, for any width $\c$, all channels will be trained evenly (2 times) by training $\c$ and its complementary together. However, this fairness can not be ensured in UA principle, but gets worse; some channels will be trained 2 times while others will be trained one or zero time.}
	\label{complementary}
	\vspace{-5mm}
\end{figure*}

\subsection{BCNet as a new supernet}

As illustrated previously, for evaluation of width $c$ at certain layer, unilaterally augmented (UA) principle assigns the leftmost $c$ channels to indicate its performance as Figure \ref{motivation}(a). Hence all channels used for width $c$ can be indexed by a set $\I_{UA}(c)$, \ie,
\begin{equation}
\I_{UA}(c) = [1:c]. 
\label{ua_channel}
\end{equation}
However, UA principle imposes an unfairness in updating channels (filters) for supernet. Channels with small index will be assigned to both small and large widths. Since different widths are sampled uniformly during the training of supernet, kernels for channels with smaller index thus get more training accordingly. To quantify this training unfairness, we can use the number of times that a channel is used for the evaluation of all widths to reflect its \textit{training degree}, and we name it as  \emph{cardinality}. Suppose a layer has maximum $l$ channels, then the cardinality for the $c$-th channel in UA principle is
\begin{equation}
\mbox{Card-UA}(c) = l-c+1.
\label{ua_counts}
\end{equation}
In this way, the cardinality of all channels varies significantly and thus they get trained much differently, which introduces evaluation bias when we use the trained supernet to rank the performance over all widths.



To alleviate the evaluation bias over widths, our proposed BCNet serves as a new supernet which promote the fairness \wrt~channels. As shown in Figure \ref{motivation}(b), in BCNet each width is simultaneously evaluated by the sub-networks corresponding to left and right channels. left and right channels. It can be seen as two identical networks $\N_l$ and $\N_r$ bilaterally coupled with each other, and use UA principle for evaluation but in a reversed order of counting channels. In this way, all channels  $\I_{BC}(c)$ used for evaluating width $c$ in BCNet are indexed by
\begin{align}
\I_{BC}(c) &= \I_{UA}^l(c) \uplus \I_{UA}^r(c) \\
&= [1:c] \uplus [(l-c+1):(l-c)],   
\label{bc_channel}
\end{align}
where $\uplus$ means the merge of two lists with repeatable elements. In detail, left channels in $\N_l$ follow the same setting with UA principle as Eq.\eqref{ua_channel}, while for right channels in $\N_r$, we count channels starting from right with $\I_{UA}^r(c) = [(l-c+1):(l-c)]$. 
As a result, the cardinality of each channel within BC principle is the sum from both two supernets $\N_l$ and $\N_r$. In detail, since channels count from the right side within $\N_r$, the cardinality for the $c$-th channel in left side corresponds to the cardinality of $l - c + 1$-th channel in right side with Eq.\eqref{ua_counts}. As a result, the cardinality for the $c$-th channel in BC principle is 
\begin{multline}
\mbox{Card-BC}(c)  = \mbox{Card-UA}(c) + \mbox{Card-UA}(l+1-c) \\
= (l-c+1) + (l+1-l-1+c) = l+1 
\label{bc_counts}
\end{multline}
Therefore, the cardinality for each channel will always amounts to the same constant value (\ie, $7$ in Figure \ref{motivation}(b)) of widths, and irrelevant with the index of channels with BC principle, thus ensuring the fairness in terms of channel (filter) level, which promotes to fairly rank network widths with our BCNet.

\subsection{Stochastic Complementary Training Strategy} 
To train the BCNet, we adopt stochastic training, \ie, uniformly sampling a network width $\c$ from the search space $\C_K$, and training its corresponding channel (filters) $\N(W,\c)$ using training data $\D_{tr}$ afterwards. Note that a single $\c$ has two paths in BCNet, during training, a training batch $\B\subset\D_{tr}$ is supposed to forward simultaneously through both $\mathcal{N}^*_l(W)$ and $\mathcal{N}^*_r(W)$. Then the training loss is the averaged loss of both paths, \ie, for each batch $\B$
\begin{equation}
\begin{aligned}
\L_{tr}(W,\c;\B) = \frac{1}{2} \cdot\left(
\L_{tr}(\mathcal{N}_l; \c, \B) + \L_{tr}(\mathcal{N}_r; \c, \B)\right).
\end{aligned}
\label{eq7}
\end{equation}	


Despite with our BCNet, channels are trained more evenly than other methods. However, it still can not ensure a rigorous fairness over channels. For example, if a layer has 3 channels and we sample 10 widths on this layer. Then results can come to that the first channel is sampled 4 times and the other two are sampled 3 times, respectively. The first channel thus still gets more training than the others, which ruins the training fairness.

To solve this issue, we propose to leverage a complementary training strategy, \ie, after sampling a network width $\c$, both $\c$ and its complementary $\bar{\c}$ get trained. For example, suppose a width $\c=(3,2,4)$ with maximum 6 channels per layer, then its complementary amounts to $\bar{\c} = (3,4,2)$ as Figure \ref{complementary}. The training loss for the BCNet is thus
\begin{equation}
\L_{tr}(W;\D_{tr},\N) = \Exp_{\bm{c}\in U(\C)} \qiLeft \mathcal{L}_{tr}(W,\bm{c};\D_{tr})  + \mathcal{L}_{tr}(W,\bar{\bm{c}};\D_{tr})\qiRight. 
\label{eq8}
\end{equation}	
In this way, when we sample a width $\c$, we can always ensure all channels are evenly trained, and expect a more fair comparison over all widths based on the trained BCNet. Note that this complementary strategy only works for our BCNet, and fails in the typical unilateral augmented (UA) principle \cite{autoslim,metapruning,tas}, which even worsens the bias as shown in Figure \ref{complementary}(b).


\subsection{Evolutionary Search with Prior Initial Population Sampling}

After the BCNet $\N^*$ is trained with weights $W^*$, we can evaluate each width by examining its performance (\eg, classification accuracy) on the validation dataset $\D_{val}$ as Eq.\eqref{eq3}. Besides, similar to the training of BCNet, the performance of a width $\c$ is indicated by the averaged accuracy of its left and right paths. Moreover, to boost the searching performance, we leverage the multi-objective NSGA-\Roman{2} \cite{deb2002fast} algorithm for evolutionary search, and hard FLOPs constraint can be thus well integrated. In generally, evolutionary search is prone to the initial population before the sequential mutation and crossover process. In this way, we propose a Prior Initial Population Sampling method to allocate a promising initial population, which is expected to contribute to the evolutionary searching performance. 





Concretely, suppose the population size is $P$, and we hope the sampled initial population have high performance in order to generate competing generations during search. Note that during training of BCNet, we have also sampled various widths, whose quality can be reflected by the training loss. In this way, we can record the top $m$ (\eg, $m=100$) widths $\{\c^{(i)}\}_{i=1}^m$ with smallest training loss $\{\ell^{(i)}\}_{i=1}^m$ as priors for good network widths. However, even the group size for every layer is set to 10, the search space of MobileNetV2 is as large as $10^{25}$, which is too large to search good widths within limited training epochs. Thus we aim to learn layer-wise discrete sampling distributions $\P(l,c_i)$ to perform stochastic sampling a width $\c = (c_1,.,c_l,.,c_L)$, where $\P(l, c_i)$ indicates the probability of sampling width $c_l$ at $l$-th layer subject to $\sum_{i}\P(l, c_i) = 1$.


Note that these $m$ prior network widths actually can reflect the preference over some widths for each layer. For example, if at a layer $l$,  a width $c_l$ exists in these $m$ prior widths with smaller training loss, then the sampling probability $\P(l,c_i)$ should be large as well. In this way, we can measure the \textit{potential error} $\E(l,c_i)$ of sampling $c_l$ width at $l$-th layer by recording the averaged training loss of all $m$ widths going through it,\ie, 
\begin{equation}
\E(l,c_i) = \frac{1}{\sum_{j=1}^m \mathbf{1}\{\c^{(j)}_l = i\}} \cdot \sum_{j=1}^m \ell^{(j)} \cdot \mathbf{1}\{\c^{(j)}_l = i\},
\label{eq10}
\end{equation}	
where $\mathbf{1}\{\cdot\}$ is the indicator function. Then the objective is to sample with minimum expected potential errors, \ie,
\begin{equation}
\begin{aligned}
\mathop{\min}_{\P}~\sum_{l}\sum_{i}&\P(l, c_i) \cdot \E(l, c_i),~\st~\sum_{i}\P(l, c_i) = 1, \\
&\P(l, c_i)\geq 0, \forall~l = 1,...,L.
\end{aligned}
\label{eq11}
\end{equation}	

\begin{figure}[t]
	\centering
	\includegraphics[width=0.7\linewidth]{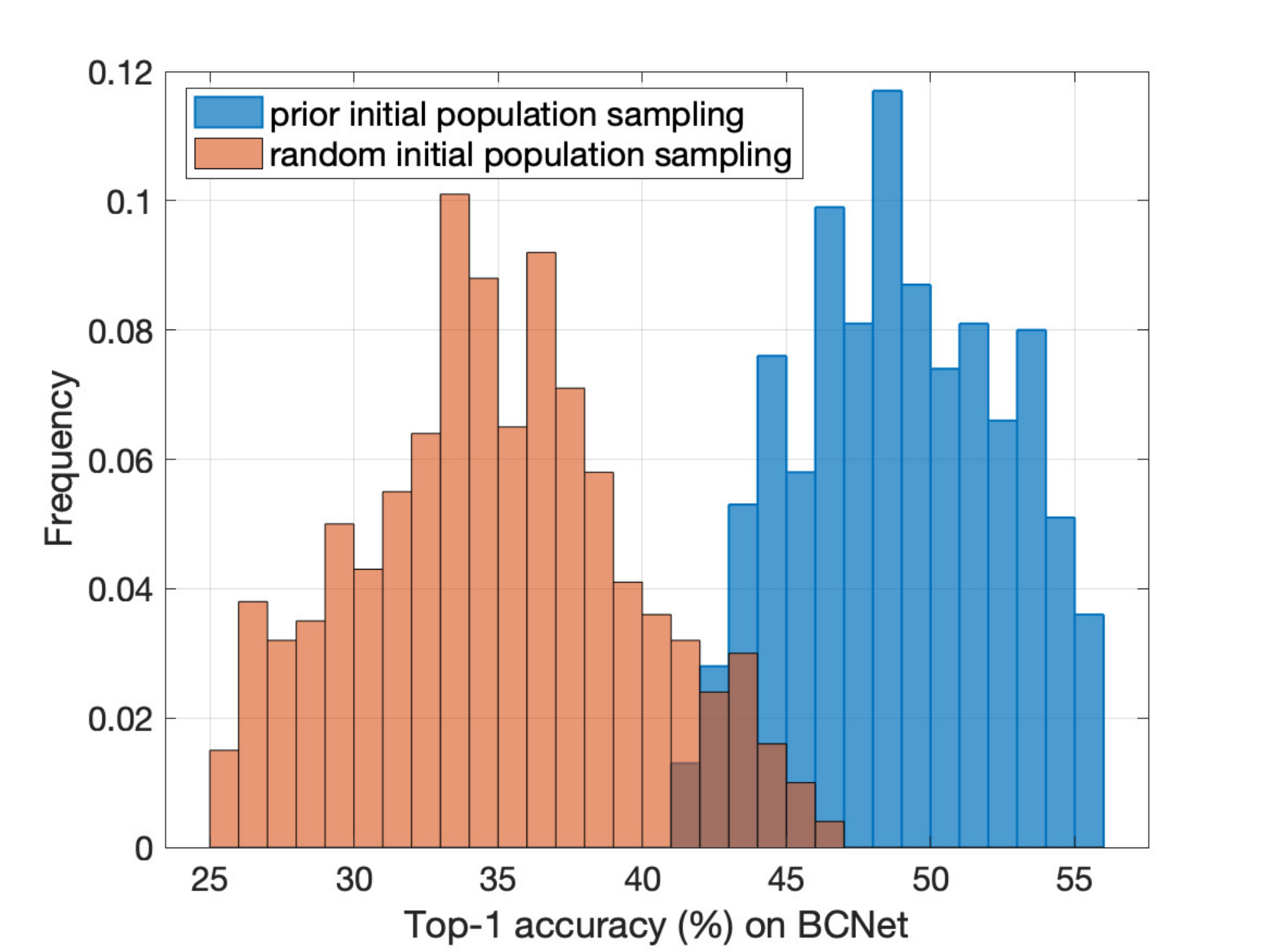}
	\caption{Histogram of Top-1 accuracy of searched widths on BCNet by evolutionary searching method with our prior or random initial population sampling \wrt~ResNet50 (2G FLOPs) on ImageNet dataset.}
	\label{prior}
	\vspace{-6mm}
\end{figure}

In addition, we also need to deal with the hard FLOPs constraint in the initial population. Since the FLOPs of a layer depends on the channels of its input and output, we can limit the expected FLOPs of the sampled network width, \ie, 
\begin{equation}
\sum_{l}\sum_{(i,j)} \P(l, c_i) \cdot F(l,c_i,c_j) \cdot \P(l+1, c_j) \leq F_b,
\label{eq12}
\end{equation}	
where $F(l,c_i,c_j)$ is the FLOPs of $l$-th layer with $c_i$ input channels and $c_j$ output channels, which can be pre-calculated and stored in a looking-up table. Then Eq.\eqref{eq12} is integrated as an additional constraint for the problem Eq.\eqref{eq11}. The overall problem is a  QCQP (Quadratically constrainted quadratic programming), which can be efficiently solved by many off-the-shelf solvers, such as CVXPY \cite{cvxpy}, GH \cite{qcqp}. As Figure \ref{prior} shows, our proposed sampling method can significantly boost evolutionary search by providing better initial populations. On average, the performance of our searched widths are much better than those obtained by random initial population, which proves the effectiveness of our proposed sampling method. 


\begin{table*}[!t]
	\centering
	\scriptsize
	\caption{Performance comparison of ResNet50 and MobileNetV2 on ImageNet. Methods with "*" denotes  that the results are reported with knowledge distillation.}
	\label{Experiments_Imagenet}
	{\begin{tabular}{c|c|cc|cc||c|c|cc|cc}
			\hline
			\multicolumn{6}{c||}{ResNet50} & \multicolumn{6}{c}{MobileNetV2}\\ \hline
			FLOPs level&Methods&FLOPs&Parameters&Top-1&Top-5&FLOPs level&Methods&FLOPs&Parameters&Top-1&Top-5  \\ \hline
			\multirow{7}*{3G} & AutoSlim* \cite{autoslim} & 3.0G & 23.1M & 76.0\% & - &  \multirow{5}*{305M (1.5$\times$)} & AutoSlim* \cite{autoslim} & 305M & 5.7M & 74.2\% & - \\
			& MetaPruning \cite{metapruning} & 3.0G & - & 76.2\% & - && Uniform & 305M & 3.6M & 72.7\% & 90.7\% \\
			& LEGR \cite{legr} & 3.0G & -  & 76.2\% & - && Random & 305M & - & 71.8\% & 90.2\% \\
			& Uniform & 3.0G & 19.1M & 75.9\% & 93.0\% && \textbf{BCNet} & 305M & 4.8M & \textbf{73.9\%} & 91.5\% \\ 
			& Random & 3.0G & - & 75.2\% & 92.5\% && \textbf{BCNet*} & 305M & 4.8M & \textbf{74.7\%} & 92.2\% \\ \cline{7-12}
			& \textbf{BCNet} & 3.0G & 22.6M & \textbf{77.3\%} & 93.7\% &\multirow{12}*{200M} & Uniform & 217M & 2.7M & 71.6\% & 89.9\% \\ \cline{1-6}
			& SSS \cite{sss} & 2.8G & - & 74.2\% & 91.9\% && Random & 217M & - & 71.1\% & 89.6\% \\
			
			\multirow{12}*{2G} & GBN \cite{gbn} & 2.4G & 31.83M & 76.2\% & 92.8\% && \textbf{BCNet} & 217M & 3.0M & \textbf{72.5\%} & 90.6\% \\
			& SFP \cite{sfp} & 2.4G & - & 74.6\% & 92.1\% && \textbf{BCNet*} & 217M & 3.0M & \textbf{73.5\%} & 91.3\% \\
			& LEGR \cite{legr} & 2.4G & - & 75.7\% & 92.7\% && MetaPruning \cite{metapruning} & 217M & - & 71.2\% & - \\
			& FPGM \cite{fpgm} & 2.4G & - & 75.6\% & 92.6\% && LEGR \cite{legr} & 210M & - & 71.4\% & - \\
			& TAS* \cite{tas} & 2.3G & - & 76.2\% & 93.1\% && AMC \cite{amc} & 211M & 2.3M & 70.8\% & - \\
			& DMCP \cite{dmcp} & 2.2G & - & 76.2\% & - && AutoSlim* \cite{autoslim} & 207M & 4.1M & 73.0\% & - \\
			& MetaPruning \cite{metapruning} & 2.0G & -
			& 75.4\% & - && Uniform & 207M & 2.7M & 71.2\% & 89.6\% \\
			& AutoSlim* \cite{autoslim} & 2.0G & 20.6M & 75.6\% & - && Random & 207M & - & 70.5\% & 89.2\% \\
			& Uniform & 2.0G & 13.3M & 75.1\% & 92.7\% && \textbf{BCNet} & 207M & 3.1M & \textbf{72.3\%} & 90.4\% \\
			& Random & 2.0G & - & 74.6\% & 92.2\% && \textbf{BCNet*} & 207M & 3.1M & \textbf{73.4\%} & 91.2\% \\ \cline{7-12}
			& \textbf{BCNet} & 2.0G & 18.4M & \textbf{76.9\%} & 93.3\% &\multirow{11}*{100M} & MetaPruning \cite{metapruning} & 105M & - & 65.0\% & - \\ \cline{1-6}
			
			\multirow{11}*{1G} & AutoPruner \cite{autopruner} & 1.4G & -  & 73.1\% & 91.3\% && Uniform & 105M & 1.5M & 65.1\% & 89.6\% \\
			& MetaPruning \cite{metapruning} & 1.0G & - & 73.4\% & - && Random & 105M & - & 63.9\% & 89.2\% \\
			& AutoSlim* \cite{autoslim} & 1.0G & 13.3M & 74.0\% & - && \textbf{BCNet} & 105M & 2.3M & \textbf{68.0\%} & 89.1\% \\
			& Uniform & 1.0G & 6.9M & 73.1\% & 91.8\% && \textbf{BCNet*} & 105M & 2.3M & \textbf{69.0\%} & 89.9\% \\
			& Random & 1.0G & - & 72.2\% & 91.4\% && MuffNet \cite{muffnet} & 50M & - & 50.3\% & - \\ 
			
			& \textbf{BCNet} & 1.0G & 12M & \textbf{75.2\%} & 92.6\% && MetaPruning \cite{metapruning} & 43M & -& 58.3\% & - \\
			
			& AutoSlim* \cite{autoslim} & 570M & 7.4M & 72.2\% & - && Uniform & 50M & 0.9M & 59.7\% & 82.0\% \\
			& Uniform & 570M & 6.9M & 71.6\% & 90.6\% && Random & 50M & - & 57.4\% & 81.2\% \\
			& Random & 570M & - & 69.4\% & 90.3\% && \textbf{BCNet} & 50M & 1.6M & \textbf{62.7\%} & 83.7\% \\
			& \textbf{BCNet} & 570M & 12.0M & \textbf{73.2\%} & 91.1\% && \textbf{BCNet*} & 50M & 1.6M & \textbf{63.8\%} & 84.6\% \\ \hline
	\end{tabular}}	
	\vspace{-5mm}
\end{table*}

\section{Experimental Results}
In this section, we conduct extensive experiments on the ImageNet and CIFAR-10 datasets to validate the effectiveness of our algorithm. For all structures, we search on the reduced space $\C_K$ with default $K=20$.  Note that most pruning methods do not report their results by incorporating the knowledge distillation (KD) \cite{distilling,du2020agree} improvement in retraining except for MobileNetV2. Thus in our method, except for MobileNetV2, we also do not include KD in final retraining. Detailed experimental settings are elaborated in supplementary materials.



\textbf{Comparison methods.} We include multiple competing pruning, network width search methods and NAS models for comparison, such as DMCP \cite{dmcp}, TAS \cite{tas}, AutoSlim \cite{autoslim}, MetaPruning \cite{metapruning}, AMC \cite{amc}, DCP \cite{dcp}, LEGR \cite{legr}, CP \cite{cp}, AutoPruner \cite{autopruner}, SSS \cite{sss}, EfficientNet-B0 \cite{efficientnet} and ProxylessNAS \cite{proxylessnas}. Moreover, we also consider two vanilla baselines. \textit{Uniform}: we shrink the width of each layer with a fixed factor to meet FLOPs budget. \textit{Random}: we randomly sample 20 networks under FLOPs constraint, and train them by 50 epochs, then we continue training the one with the highest performance and report its final result.


\begin{table}[!t]
	\centering
	\footnotesize
	\caption{Searching results of EfficientNet-B0 and ProxylessNAS on ImageNet dataset.}
	\label{Experiments_Efficientnetb0}
	{\begin{tabular}{c|c|c|c|c}
			\hline
			\multicolumn{5}{c}{EfficientNet-B0} \\ \hline
			Groups&Methods&Param&Top-1&Top-5  \\ \hline
			\multirow{3}*{385M}&Uniform & 5.3M & 76.88\% & 92.64\% \\
			&Random & 5.1M & 76.37\% & 92.25\% \\
			& BCNet & 6.9M & \textbf{77.36\%} & 93.17\% \\ \hline
			\multirow{3}*{192M}&Uniform & 2.7M & 74.26\% & 92.24\% \\
			&Random & 2.9M & 73.82\% & 91.86\% \\
			& BCNet & 3.8M & \textbf{74.92\%} & 92.06\% \\ \hline
			\multicolumn{5}{c}{ProxylessNAS} \\ \hline
			Groups&Methods&Param&Top-1&Top-5  \\ \hline
			\multirow{3}*{320M} & Uniform & 4.1M & 74.62\% & 91.78\% \\
			& Random & 4.3M & 74.16\% & 91.23\% \\
			&BCNet & 5.4M & \textbf{75.07\%} & 91.97\% \\ \hline
			\multirow{3}*{160M} & Uniform & 2.2M & 71.16\% & 89.49\% \\
			&Random & 2.5M & 70.89\% & 89.12\% \\
			&BCNet & 2.9M & \textbf{71.87\%} & 89.96\% \\ \hline
	\end{tabular}}	
	\vspace{-5mm}
\end{table}

\subsection{Results on ImageNet Dataset} 
ImageNet dataset contains 1.28M training images and 50K validation images from 1K classes. In specific, we report the accuracy on the validation dataset as \cite{slimming,autoslim}, and the original model takes as the supernet while for the 1.0$\times$ FLOPs of all models, the supernet refers to a 1.5$\times$ FLOPs of original model by uniform width scaling. To verify the performance on both heavy and light models, we search on the ResNet50 and MobileNetV2 with different FLOPs budgets. In our experiment, the original ResNet50 (MobileNetV2) has 25.5M (3.5M) parameters and 4.1G (300M) FLOPs with 77.5\% (72.6\%) Top-1 accuracy, respectively.

As shown in Table \ref{Experiments_Imagenet}, our BCNet achieves the highest accuracy on ResNet50 and MobileNetV2 \wrt~different FLOPs, which indicates the superiority of our BCNet to other pruning methods. For example, our 3G FLOPs ResNet50 decreases only 0.2\% Top-1 accuracy compared to the original model, which exceeds AutoSlim \cite{autoslim} and MetaPruning \cite{metapruning} by 1.3\% and 1.1\%. While for MobileNetV2, our 207M MobileNetV2 exceeds the state-of-the-art AutoSlim, MetaPruning by 0.4\%, 1.1\%, respectively. In addition, our BCNet even surpasses other algorithms more on tiny MobileNetV2 (105M) with 68\% Top-1 accuracy and exceeds MetaPruning by 3.0\%.


To further demonstrate the effectiveness of our BCNet on highly efficient models, we conduct searching on the NAS-based models EfficientNet-B0 and ProxylessNAS. The original EfficientNet-B0 (ProxylessNAS) has 5.3M (4.1M) parameters and 385M (320M) FLOPs with 76.88\% (74.62\%) Top-1 accuracy, respectively. As shown in Table \ref{Experiments_Efficientnetb0}, although the increase of performance is not as signicant as in Table \ref{Experiments_Imagenet}, our method can still boost the NAS-based models by more than 0.4\% on Top-1 accuracy.

\begin{table}[t]
	\centering
	\caption{Performance comparison of MobileNetV2 and VGGNet on CIFAR-10.}
	\label{Experiment_Cifar10}
	\scriptsize
	{\begin{tabular}{c|c|ccc}
			\hline
			\multicolumn{5}{c}{MobileNetV2} \\ \hline
			Groups&Methods&FLOPs&Params&accuracy  \\ \hline
			\multirow{4}*{200M} & DCP \cite{dcp} & 218M & - & 94.69\% \\
			& Uniform & 200M & 1.5M & 94.57\% \\
			& Random & 200M & - & 94.20\% \\
			& \textbf{BCNet} & 200M & 1.5M & \textbf{95.44\%}\\ \cline{1-5}
			
			\multirow{4}*{146M} & MuffNet \cite{muffnet} & 175M & - & 94.71\% \\ 
			& Uniform & 146M & 1.1M & 94.32\% \\
			& Random & 146M & - & 93.85\% \\
			& \textbf{BCNet} & 146M & 1.2M & \textbf{95.42\%} \\ \cline{1-5}
			
			\multirow{5}*{44M} & AutoSlim \cite{autoslim} & 88M & 1.5M & 93.20\% \\
			& AutoSlim \cite{autoslim} & 59M & 0.7M & 93.00\% \\
			& MuffNet \cite{muffnet} & 45M & - & 93.12\% \\
			& Uniform & 44M & 0.3M & 92.88\% \\
			& Random & 44M & - & 92.31\% \\
			& \textbf{BCNet} & 44M  & 0.4M & \textbf{94.42\%} \\ \cline{1-5}
			
			\multirow{4}*{28M} & AutoSlim \cite{autoslim} & 28M & 0.3M & 92.00\%\\
			& Uniform & 28M & 0.2M & 92.37\% \\
			& Random & 28M & - & 91.69\% \\
			& \textbf{BCNet} & 28M & 0.2M & \textbf{94.02\%} \\ \hline
			\multicolumn{5}{c}{VGGNet} \\ \hline
			Groups&Methods&FLOPs&Params&accuracy  \\ \hline
			\multirow{5}*{200M} & Sliming \cite{slimming} & 199M & 10.4M & 93.80\% \\
			& DCP \cite{dcp} & 199M & 10.4M & 94.16\% \\
			& Uniform & 199M & 10.0M & 93.45\%  \\
			& Random & 199M & - & 93.02\% \\ \cline{1-5}
			& \textbf{BCNet} & 197M & 3.1M & \textbf{94.36\%}  \\ 
			\multirow{8}*{100M$+$} & Uniform & 185M & 9.3M & 93.30\% \\
			& Random & 185M & - & 92.71\%  \\
			& \textbf{BCNet} & 185M & 6.7M & \textbf{94.14\%} \\ 
			& CP \cite{cp} & 156M & 7.7M & 93.67\%  \\
			& Multi-loss \cite{multi} & 140M & 5.5M & 94.05\% \\
			& Uniform & 138M & 6.8M & 93.14\% \\
			& Random & 138M & - & 92.17\% \\
			& \textbf{BCNet} &  138M & 3.3M & \textbf{94.09\%} \\ \cline{1-5}
			\multirow{5}*{77M} & CGNets \cite{cgnet} & 91.8M & - & 92.88\% \\
			& Uniform & 77.0M & 3.9M & 92.38\% \\
			& Random & 77.0M & - & 91.72\%  \\
			& \textbf{BCNet} & 77.0M & 1.2M & \textbf{93.53\%} \\
			& CGNet \cite{cgnet} & 61.4M & - & 92.41\% \\ \hline
	\end{tabular}}	
	\vspace{-7mm}
\end{table}

\subsection{Results on CIFAR-10 Dataset}
We also examine the performance of MobileNetV2 and VGGNet on the moderate CIFAR-10 dataset, which has 50K training and 10K testing images with size 32$\times$32 of 10 categories. Our original VGGNet (MobileNetV2) has 20M (2.2M) parameters and 399M (297M) FLOPs with accuracy of 93.99\% (94.81\%).

As shown in Table \ref{Experiment_Cifar10}, our BCNet still enjoys great advantages in various FLOPs levels. For instance, our 200M MobileNetV2 can achieve 95.44\% accuracy, which even outperforms the original model by 0.63\%. Moreover, even with super tiny size (28M), our BCNet can still have 94.02\% accuracy, which surpasses the state-of-the-art AutoSlim\cite{autoslim} by 2.0\%. As for VGGNet, our BCNet is capable of outperforming those competing channel pruning methods DCP \cite{dcp} and Slimming \cite{slimming} by 0.20\% and  0.56\% with 2$\times$ acceleration rate.


\subsection{Ablation Studies}
\label{ablation}

\textbf{Effect of BCNet as a supernet.} To validate the effectiveness of our proposed supernet BCNet, we search the ResNet50, MobileNetV2, EfficientNet-B0 and ProxylessNAS on ImageNet dataset with 2$\times$ acceleration. Our default baseline supernet is that adopted by AutoSlim \cite{autoslim}, which follows unilateral augmented principle to evaluate a network width. As the results in Table \ref{BCNet_analysis} shows, under the greedy search, only using our BCNet evaluation mechanism (second line) can enjoy a gain of 0.27\% to 0.66\% Top-1 accuracy. When searching with evolutionary algorithms, the gain still reaches at 0.28\% to 0.35\% Top-1 accuracy on various models. These exactly indicates using BCNet as supernet could boost the evaluation and searching performance. As for the complementary training strategy, we can see that it enables to boost our BCNet by improving the MobileNetV2 (ResNet50) from 69.92\% (76.41\%) to 70.04\% (76.56\%) on Top-1 accuracy. Note that greedy search without BCNet supernet amounts to AutoSlim, we can further indicate the superiority of our method to AutoSlim with achieved Top-1 accuracy 70.20\% (76.90\%) vs 69.52\% (75.94\%) on MobileNetV2 (ResNet50).

\begin{table*}[t]
	\caption{Performance of searched MobileNetV2 (150M FLOPs), ResNet50 (2G FLOPs), EfficientNet-B0 and ProxylessNAS on ImageNet dataset with different supernet and searching methods.}
	\label{BCNet_analysis}
	\centering
	\scriptsize
	\begin{tabular}{|c|c|c|c|c||cc|cc|cc|cc|} \hline
		\multicolumn{2}{|c|}{evaluator} & \multicolumn{3}{c||}{searching}&  \multicolumn{8}{c|}{models} \\ \cline{1-13} 
		BCNet & complementary & greedy & \multicolumn{2}{c||}{evolutionary}& \multicolumn{2}{c|}{MobileNetV2} & \multicolumn{2}{c|}{ResNet50}&\multicolumn{2}{c|}{EfficientNet-B0}& \multicolumn{2}{c|}{ProxylessNAS}\\ \cline{4-13} 
		supernet &training&search&random & prior & Top-1 & Top-5&Top-1 & Top-5&Top-1 & Top-5&Top-1 & Top-5\\ \hline 
		& &\checkmark& & & 69.52\% & 88.91\% & 75.64\% & 92.90\% & 74.02\% & 91.58\% & 70.97\% & 89.43\% \\
		\checkmark & &\checkmark& & & 69.87\% & 88.99\% & 76.30\% & 93.16\% & 74.39\% & 91.66\% & 71.24\% & 89.57\%  \\
		\checkmark & \checkmark &\checkmark& & & \textbf{69.91\%} & 89.02\% & \textbf{76.42\%} & 93.19\% & \textbf{74.51\%} & 91.78\% & \textbf{71.33\%} & 89.62\%\\ \hline
		& & & \checkmark &  & 69.64\% & 88.85\% & 76.12\% & 92.95\% & 74.35\% & 91.54\% & 71.13\% &  89.49\% \\
		\checkmark & & & \checkmark &  & 69.92\% & 88.91\% & 76.41\% & 93.12\% & 74.63\% & 91.93\% & 71.48\% & 89.69\%  \\
		\checkmark & \checkmark & & \checkmark &  & 70.04\% & 89.02\% & 76.56\% & 93.21\% & 74.73\% & 91.85\% & 71.62\% & 89.73\% \\
		\checkmark & \checkmark & &  & \checkmark & \textbf{70.20\%} & 89.10\% & \textbf{76.90\%} & 93.30\% & \textbf{74.92\%} & 92.06\% & \textbf{71.87\%} & 89.96\% \\ \hline
	\end{tabular} 
	\vspace{-4mm}
\end{table*}

\begin{figure}[t]
	\centering
	\includegraphics[width=0.71\linewidth]{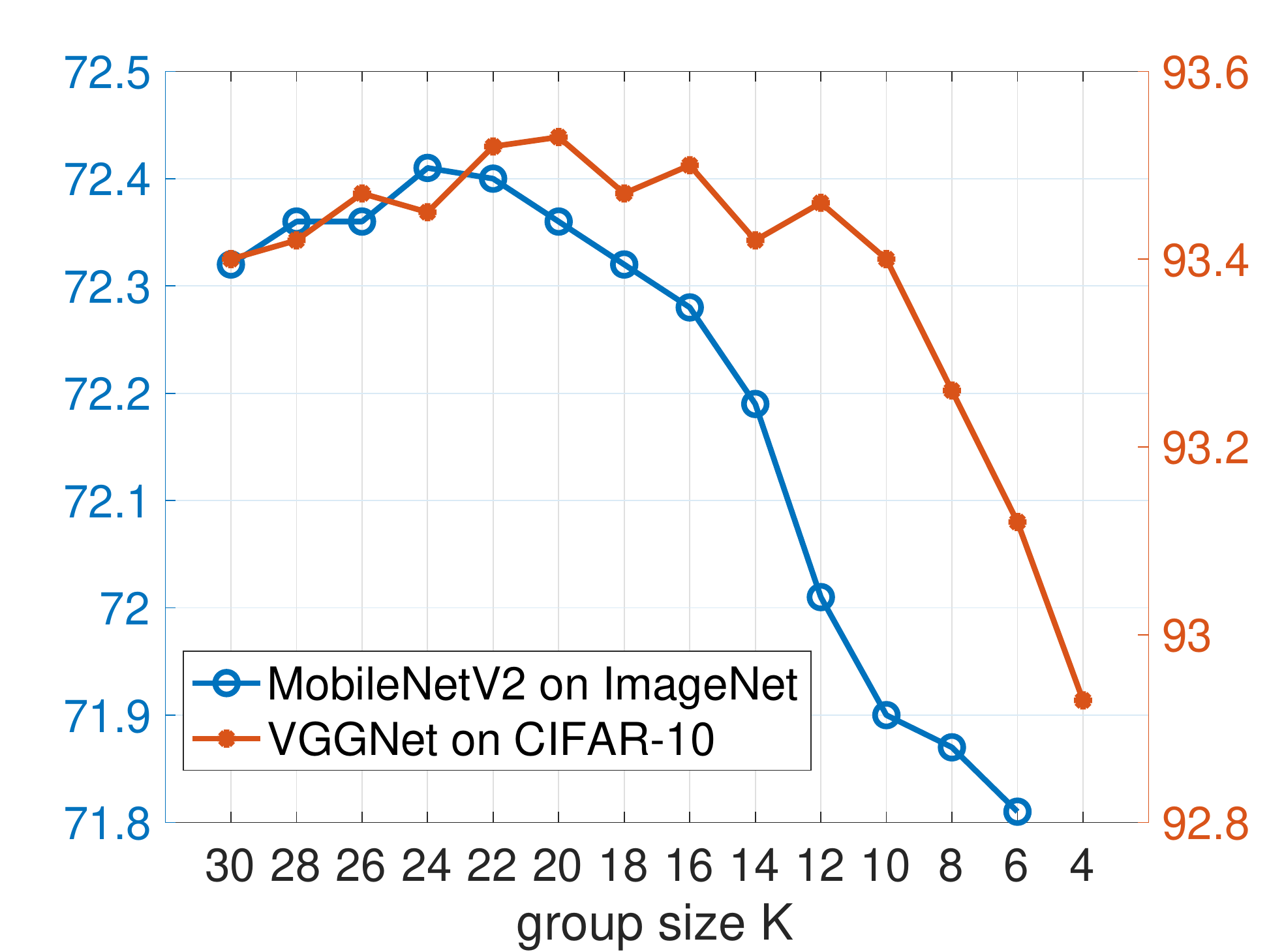}
	\vspace{-1mm}
	\caption{Accuracy performance of the searched network with different group size $K$ of the search space.}
	\label{ratio_list}
	\vspace{-6mm}
\end{figure}

\begin{figure*}[t]
	\centering
	\includegraphics[width=1.00\linewidth]{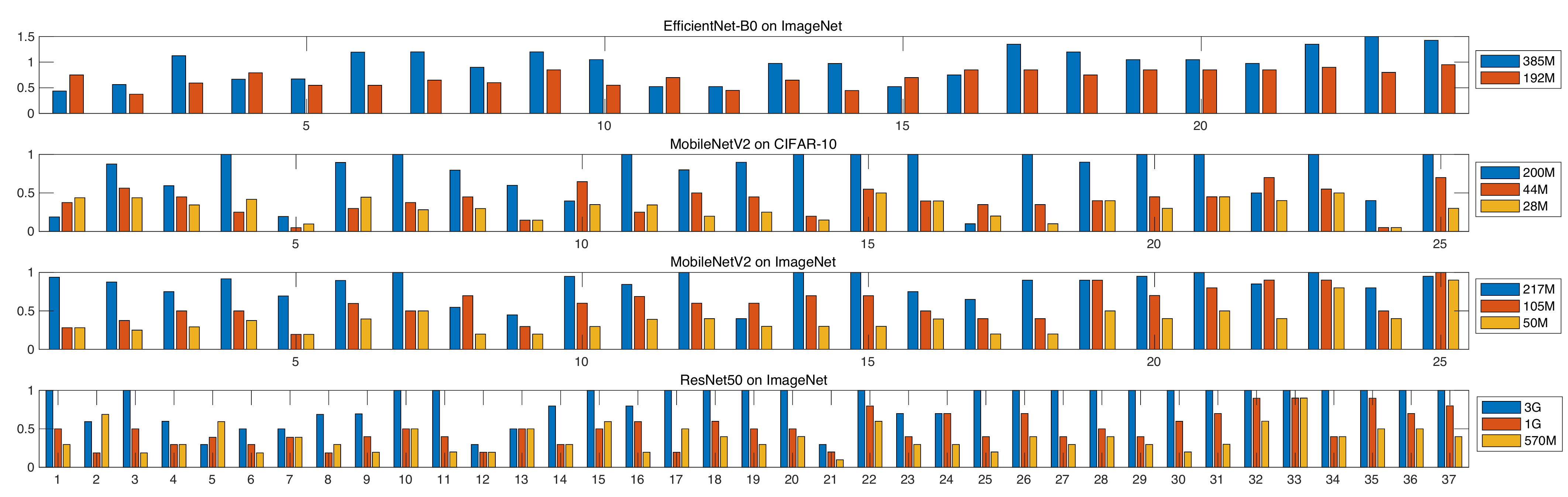}
	\vspace{-6mm}
	\caption{Visualization of searched networks \wrt~different FLOPs. The vertical axis means the ratio of retained channel number compared to that of original networks at each layer.}
	\label{visualization}
	\vspace{-5mm}
\end{figure*}

\textbf{Effect of search space.}
We adopt the grouped search space $\C_K$ to reduce its complexity. To investigate the effect of search space, we searched VGGNet on CIFAR-10 dataset and MobileNetV2 on ImageNet dataset with various group size $K$. As in Figure \ref{ratio_list}(b), our method achieves the best performance in most cases around our default value $K=20$. In addition, we noticed that when the group size $K$ is small, the performance of searched network will increase with $K$ growing larger. This is because group size $K$ determines the size of search space $\C_K$, and larger $K$ induces larger search space. In this way, the obtained network width will be closer to the Oracle optimal width, with higher accuracy achieved accordingly. In addition, the performance tends to be stable when the group size lies in $[14:22]$ but decreases afterwards, which implies the searching space might be  too large for searching an optimal width.

\subsection{Visualization and Interpretation of Results}
For intuitively understanding, we visualize our searched networks with various FLOPs in Figure \ref{visualization}. Moreover, for clarity we show the retained ratio of layer widths compared to that of the original models. Note that for MobileNetV2, ResNet50, EfficientNet-B0 and ProxylessNAS with skipping or depthwise layers, we merge these layers which are required to have the same width. 
More visualization results can refer to the supplementary materials.

From Figure \ref{visualization}, we can see that on the whole, with decreasing FLOPs, layer width nearer the input tends to be reduced. However, the last layer is more likely to be retained. This might result from that the last layer is more sensitive to the classification performance, thus it is safely kept when the FLOPs is reduced. In the sequel, we will illustrate more elaborate observations \wrt~each network, and present some intuitions accordingly.

\textbf{ResNet50 on ImageNet.} We found that when the network is pruned with a large FLOPs budget (\eg, 3G or 2G), width of the first 1$\times$1 convolutional layer (\eg, $2$nd and $5$th layer in Figure \ref{visualization}) of each block in ResNet50 is preferentially reduced, which means 1$\times$1 convolution may contribute less to classification performance. However, when FLOPs drops to a fairly small value (\eg, 570M), channel number of 3$\times$3 convolution (\eg, $3$rd and $6$th layer in Figure \ref{visualization}) will decrease dramatically while that of 1$\times$1 convolution increases instead. This implies that the network will be forced to use more 1$\times$1 convolutions instead of 3$\times$3 convolutions to extract information from feature maps. In addition, this observation also indicates that evolutionary algorithm is more effective than greedy search \wrt~small FLOPs since evolutionary algorithm can always maintain the original search space. Nevertheless, greedy algorithm will greedily prune out more 1$\times$1 convolutions at the beginning, which can not be recovered for small FLOPs budget.

\textbf{MobileNetV2 on ImageNet and CIFAR-10.} Different from ResNet50, widths of MobileNetV2 decrease more evenly with the reduction of FLOPs. This may be due to the limitation of depthwise convolutions, which requires the output channel number of first 1$\times$1 convolution and the second 3$\times$3 convolution to be the same in MobileNetV2 blocks. Compared from pruning on ImageNet, widths closer to the input layer are more easily to be clipped on CIFAR-10 dataset. This may be because the input of CIFAR-10 is 32$\times$32, which do not need as many widths as ImageNet in the last layer. In addition, when FLOPs is reduced to a fairly small value (\eg, 28M, 44M, and 50M for MobileNetV2), unlike pruning on ImageNet, the width of the last layer of MobileNetV2 on CIFAR-10  decreases rapidly. The reason for this phenomenon may be that MobileNetV2 on ImageNet is forced to classify 1000 categories, while it only needs to deal with 10-way classification on CIFAR-10. Then the width of the last layer on ImageNet tends to be retained, but gets decreased rapidly on CIFAR-10. More visualizations about MobileNetV2 are analyzed in the supplementary material.

\textbf{EfficientNet-B0 on ImageNet.} EfficientNet-B0 shares similar block structure with MobileNetV2. However, the width of EfficientNet-B0 varies more evenly than MobileNetV2, which may be due to its width setting is more better since it is determined by NAS. In detail, compared to the original setting of EfficientNet-B0, for the searched 1$\times$ FLOPs network, the channels of adjacent blocks show opposite fluctuations (\eg, channels of 1,3,5 blocks increase while channels in 2,4,6 blocks decrease). This may mean that the fluctuations of widths are conducive to the performance of searched network structure.

\section{Conclusion}

In this paper, we introduce a new supernet called BCNet to address the training unfairness and corresponding evaluation bias for searching optimal network width. In our BCNet, each channel is fairly trained and responsible for the same amount of widths. Besides, we leverage a stochastic complementary strategy for the training of BCNet, and propose a prior initial population sampling method to boost the evolutionary search. Extensive experiments have been implemented on benchmark ImageNet and CIFAR-10 datasets to show the superiority of our proposed method to other state-of-the-art channel pruning/network width search methods.

\subsubsection*{Acknowledgments}
This work is funded by the National Key Research and Development Program of China (No. 2018AAA0100701) and the NSFC 61876095. Chang Xu was supported in part by the Australian Research Council under Projects DE180101438 and DP210101859.	

{\small
\bibliographystyle{ieee_fullname}
\bibliography{Reference}
}

\newpage

\onecolumn

\appendix
The supplementary materials are organized as follows. In Appendix \ref{A1}, we illustrate the details of experimental settings. In Appendix \ref{A2}, we provide the implementation details of evolutionary search. We report the effect of performance improvement with more training and searching cost in Appendix \ref{A3}. Then
we investigate the effect of the prior initial population in Appendix \ref{A4}. We conduct the performance comparison of BCNet and AutoSlim \cite{autoslim} with the same training recipe in Appendix \ref{A5}. In Appendix \ref{A6}, we investigate the effect of training BCNet with different epochs. We report more detailed searching results of BCNet on ImageNet dataset. In Appendix \ref{A7}, we include more detailed searching results for Table \ref{Experiments_Imagenet}. Then we present the performance of searched models \wrt different FLOPs in Appendix \ref{A8}. Finally, we show the visualization of searched network widths with 2$\times$ acceleration in Appendix \ref{A9}.

\section{Details of Training Recipe} \label{A1}
In this section, we present the training details of our BCNet \wrt~experiments on various
models. In detail, we search on the reduced space $\C_K$ with default $K=20$. During training, except for EfficientNet-B0 and ProxylessNAS, we use SGD optimizer with momentum 0.9 and nesterov acceleration. As for EfficientNet-B0 and ProxylessNAS, we adopt RMSprop optimizer for searching optimal network width. 

\textbf{Training Recipe of ResNet50, MobileNetV2, and VGGNet.} For ResNet50, we follow the same training recipe as TAS \cite{tas}. In detail, we use a weight decay of $10^{-4}$ and batch size of 256; and we train the model by 120 epochs with the learning rate annealed with cosine strategy from initial value 0.1 to $10^{-5}$. For MobileNetV2 and VGGNet, we set weight decay to $5\times10^{-5}$ and $10^{-4}$, respectively. Besides, for MobileNetV2, we adopt the batch size of 256, and the learning rate is annealed with a cosine strategy from initial value $0.1$ to $10^{-5}$. For VGGNet, we train it for 400 epochs using a batch size of 128; the learning rate is initialized to 0.1 and divided by 10 at 160-th, 240-th epoch. Moreover, we note that most pruning methods do not report their results by incorporating the knowledge distillation (KD) \cite{distilling} improvement in retraining except for MobileNetV2. Thus in our method, except for MobileNetV2, we do not include KD in final retraining for a more fair comparison of performance. All experiments are implemented with PyTorch on NVIDIA V100 GPUs.

\textbf{Training Recipe of EfficientNet-B0 and ProxylessNAS.} We use the same training strategies for both EfficientNet-B0 and ProxylessNAS. In detail, we train both models for 300 epochs using a batch size of 1024; the learning rate is initialized to 0.128 and decayed by 0.963 for every 3 epochs. Besides, the first 5 training epochs are adopted as warm-up epochs, and the weight decay is set to 1$\times 10^{-5}$.

\section{Details of Evolutionary Search} \label{A2}
During our evolutionary search, each network width $\c$ is indicated by the averaged accuracy of its left and right paths, which can be formulated as Eq. \eqref{supp_evolution}. The optimal width (rather than sub-nets) refers to the one with the highest performance and we then train it from scratch.
\begin{equation}
\begin{aligned}
\mbox{Accuracy}(W,\c;\D_{val}) = \frac{1}{2} \cdot\left(
\mbox{Accuracy}(\mathcal{N}_l, \c; \D_{val}) + \mbox{Accuracy}(\mathcal{N}_r, \c; \D_{val})\right).
\end{aligned}
\label{supp_evolution}
\end{equation}	
Concretely, we adopt the multi-objective NSGA-\Roman{2} \cite{deb2002fast} algorithm to implement the search. Note that some networks (\eg, MobileNetV2) may have batch normalization (BN) layers, and due to the varying network widths, the mean and variance in the BN layers are not suitable to all widths. In this way, we simply use the mean and variance in batches instead, and we set the batch size to 2048 during testing to ensure accurate estimates of the mean and variance.

In detail, we set the population size as 40 and the maximum iteration as 50. Firstly, we apply our proposed prior initial population sampling method Eq. \eqref{eq10} $\sim$ Eq.\eqref{eq12} to generate the initial population. In each iteration,  we use the trained BCNet to evaluate each width and rank all widths in the population. After the ranking, we use the tournament selection algorithm to select 10 widths retained in each generation. And the population for the next iteration is generated by two-point crossover and polynomial mutation. Finally, the searched width refers to the one with the best performance in the last iteration, and we train it from scratch for evaluation. 

\textbf{Pipeline of training and evolutionary search:} We follow a routine pipeline in width searching methods, such as TAS and AutoSlim. We first train a supernet (\ie, BCNet) and then use it to search for the optimal width by evolutionary algorithms. For each sampled width during the search, we evaluate it by the inference with the weights from BCNet and record its accuracy; since there is no network training during evolutionary, it is thus very efficient. Finally, we only retrain the width with the highest accuracy from scratch and report its performance.

\section{Effect of Performance Improvement with more Training and Searching Cost} \label{A3}
Since BCNet needs 2$\times$ of training and searching cost of the UA principle,  one intuitive question comes to \textit{whether UA principle can benefit from more training and searching cost and even surpass BCNet as a result.} With this aim, we search with UA principle on more times (1$\times$ and 2$\times$) of training epochs and iterations with evolutionary search to search for 0.5$\times$ FLOPs ResNet50 and MobileNetV2 on ImageNet dataset.
\begin{table*}[h]
	\centering
	\small
	\caption{Performance with more training and searching cost on ImageNet dataset. Note BCNet achieves 76.90\% and 70.20\% accuracy for 0.5$\times$ FLOPs ResNet50 and MobileNetV2, respectively.}
	\label{time_efficiency}
	\begin{tabular}{c|c|c||c|c|c}
		\hline
		\multicolumn{3}{c||}{ResNet50} & \multicolumn{3}{c}{MobileNetV2} \\ \hline
		\diagbox{Searching}{Training} &1$\times$&2$\times$&\diagbox{Searching}{Training} &1$\times$&2$\times$\\ \hline
		1$\times$&75.63\%& 75.65\% & 1$\times$& 68.95\% & 68.98\% \\ \hline
		2$\times$&75.74\%& 75.72\% &2$\times$& 69.03\% & 69.06\% \\ \hline
	\end{tabular}	
	\label{supp_budget}
\end{table*} 

From Table \ref{supp_budget}, we can know that evolutionary search only benefits a little from more searching iterations. Simultaneously, $2$ times of training epochs for supernet nearly does not affect the search result. As a result, our BCNet can efficiently boost the performance of searching results with two times of training and searching cost. Note that after the search, we train the searched network width from scratch for evaluation, which amounts to the same cost as other methods \cite{tas,autoslim,metapruning}.

\section{Effect of Prior Initial Population Sampling (PIPS)} \label{A4}

\begin{figure}[h]
	\centering
	\includegraphics[width=0.48\linewidth]{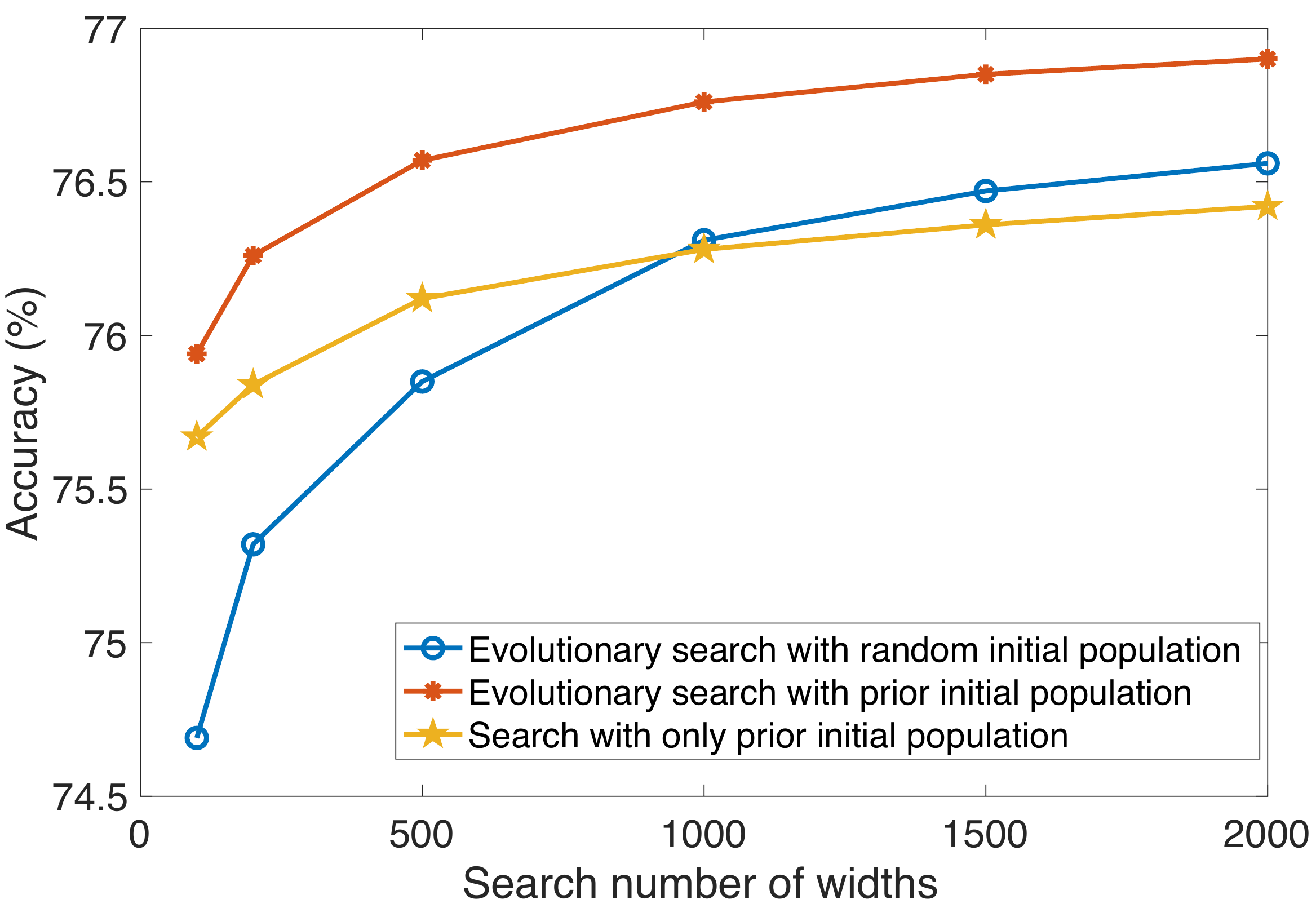}
	\caption{Top-1 accuracy of searched models on ImageNet dataset by different methods with the increasing of search numbers. }
	\label{supp_prior}
\end{figure}

Our proposed prior initial population sampling (PIPS) method aims to provide a better initial population for evolutionary search, and the searched optimal width will have higher performance accordingly. Now we want to investigate how the effect of directly leveraging PIPS to search for optimal width. With this aim, we pick up the optimal width with the highest validation Top-1 accuracy after $\{100, 200, 500, 1000, 1500, 2000\}$ of search number of widths, respectively. Then we train them from scratch and report their Top-1 accuracy in Figure \ref{supp_prior}. The search is implemented on ResNet50 on ImageNet dataset with 3 different settings and 0.5$\times$ FLOPs budget, \ie, evolutionary search with random initial population, evolutionary search with prior initial population, and search with the only prior initial population.

From Figure \ref{supp_prior}, we can know that widths provided by our prior initial population sampling method can surpass those from the random initial population by a larger gap on Top-1 accuracy. Besides, it also should be noticed that evolutionary search benefits more from the increase of search numbers, which indicates that evolutionary search can better utilize the searched width to achieve the optimal result. In addition, with our prior initial population, evolutionary search can get better network width with higher performance (\ie, \red{red} line in Figure \ref{sup_prior}), which means that our prior initial population sampling method can provide good initialization for evolutionary algorithm.  Moreover, the Top-1 accuracy of searched models rises slowly after 1000 search numbers, which may imply that the evolutionary algorithm can already find a good solution in this case.

\section{Comparison of BCNet and AutoSlim \cite{autoslim} under 305M FLOPs} \label{A5}
\begin{figure}[h]
	\centering
	\includegraphics[width=\linewidth]{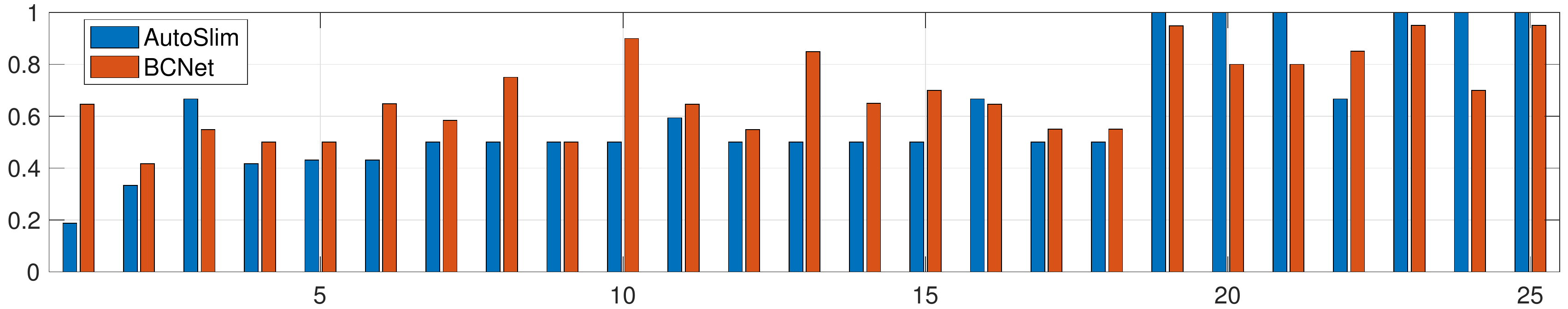}
	\caption{Visualization of searched MobileNetV2 with 305M FLOPs on ImageNet dataset. Both networks are searched on 1.5$\times$ search space as AutoSlim \cite{autoslim}.}
	\label{Compare_AutoSlim}
\end{figure}

To intuitively check the effect of BCNet with another baseline method, we visualize the network width searched by BCNet and the released structure of AutoSlim \cite{autoslim} for 305M-FLOPs MobileNetV2 in Figure \ref{Compare_AutoSlim}. In detail, BCNet saves
more layer widths in the first few layers, and prunes a bit more widths in the last few layers, which is more evenly than AutoSlim.

To promote the fair comparison of BCNet and AutoSlim, we retrain the released structure of AutoSlim (\ie, 305 FLOPs MobileNetV2) with the same training recipe of ours. Note that we do not include KD for a more fair comparison of AutoSlim \cite{autoslim}, as shown in Table \ref{Experiments_supp_autoslim}.
\begin{table*}[h]
	\centering
	\small
	\caption{Performance comparison with AutoSlim \cite{autoslim} of 305M MobileNetV2 on ImageNet by the same training recipe.}
	\label{Experiments_supp_autoslim}
	{\begin{tabular}{c|c|cc|cc}
			\hline
			Methods&FLOPs&Parameters&Top-1&Top-5 \\ \hline
			AutoSlim&305M&5.8M&73.1\%&91.1\% \\
			BCNet&305M&4.8M&73.9\%&92.2\% \\ \hline
	\end{tabular}}	
\end{table*}

From Table \ref{Experiments_supp_autoslim} and Figure \ref{Compare_AutoSlim}, we can know that BCNet retains more widths closer to the input layer, and thus the parameters of our searched structures are lesser than AutoSlim. Moreover, with the same training recipe, BCNet achieves a 0.8\% higher on Top-1 accuracy than AutoSlim with 305M FLOPs MobileNetV2, which indicates the effectiveness of our method.

\section{Transferability of the Searched Width to Object Detection Task} \label{Ax}
For object detection tasks, a pretrained model on ImageNet dataset is usually leveraged as its backbone. As a result, We take the searched 2G FLOPs ResNet50 as the backbone in detection to examine the transferability of BCNet for other tasks \cite{fpn,frcnn}. The results are reported in Table \ref{detection} for both Faster R-CNN with FPN and RetinaNet, indicating the backbones obtained by BCNet(0.5$\times$) can achieve higher performance to the uniform baseline(0.5$\times$)

\begin{table}[h]
	\centering
\caption{Detection performance with ResNet50 as the backbone.}
\label{detection}
\begin{tabular}{l|c|c|c}
		\hline
		Framework&Original (4G)&BCNet (2G)&Uniform(2G)  \\ \hline
		RetinaNet  & 36.4\% & 35.4\% & 34.3\% \\
		Faster R-CNN & 37.3\% & 36.3\% & 35.4\% \\ \hline
\end{tabular}
\end{table}

\newpage
\section{Effect of Training BCNet with Different Epochs} \label{A6}
\begin{figure}[h]
	\centering
	\includegraphics[width=0.42\linewidth]{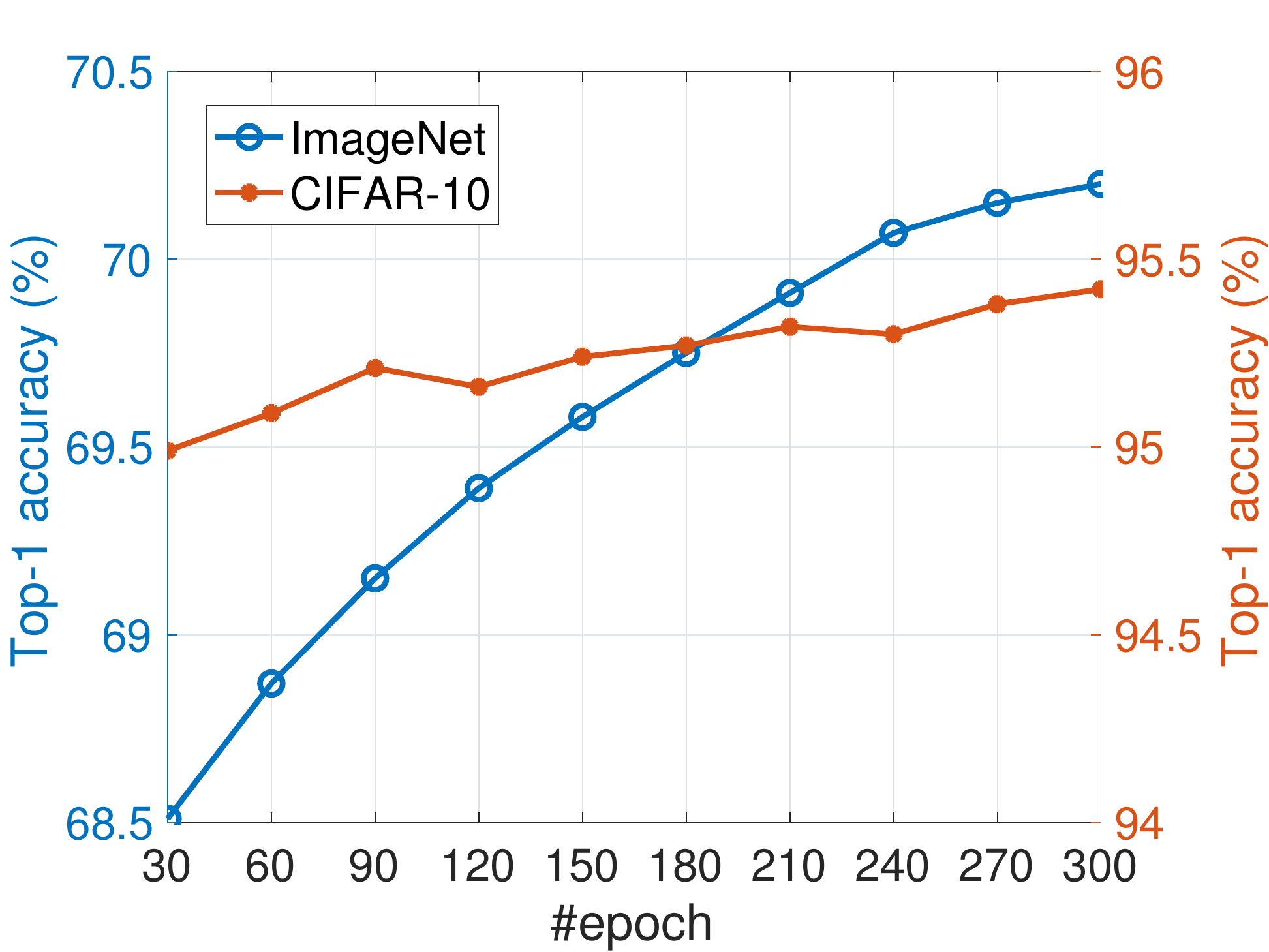}
	\caption{Performance of searched MobileNetV2 (50\% FLOPs) \wrt~different training epochs of BCNet on ImageNet and CIFAR-10 dataset.}
	\label{sup_prior}
\end{figure}

Since BCNet is critical in searching width as a fundamental performance estimator, we investigate how much effort should be spared in its training to ensure the searching performance. In this way, we train the BCNet with a different number of epochs and use these BCNets to search MobileNetV2 with 50\% FLOPs on ImageNet and CIFAR-10 datasets. The Top-1 accuracies of searched networks are presented in Figure \ref{sup_prior}.

From Figure \ref{sup_prior}, we can see that more challenging (\eg, ImageNet) datasets usually need more training epochs for the performance estimator than simple datasets (\eg, CIFAR-10). Concretely, with 30 (300) training epochs of BCNet, the searched MobileNetV2 can achieve 68.51\% (70.20\%) Top-1 accuracy on ImageNet dataset. In contrast, for CIFAR-10 dataset, the Top-1 accuracy \wrt~300 training epoch is merely 0.43\% better than that of 30 training epochs.

\section{Searching Performance with Different FLOPs} \label{A8}
\begin{figure}[h]
	\centering
	\subfigure[ImageNet.]{
		\label{supp3a}
		\includegraphics[width=0.40\linewidth]{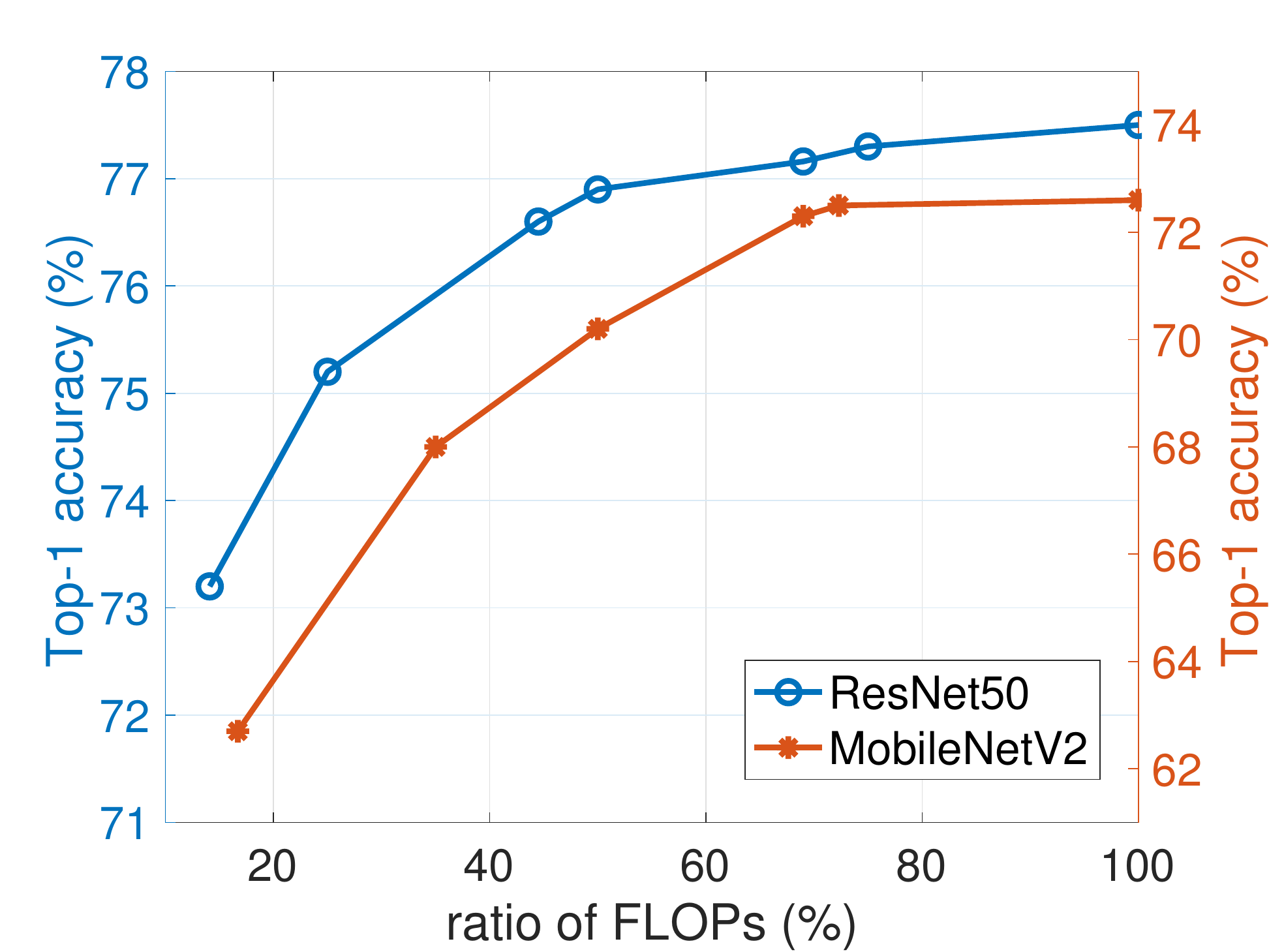}}~~
	\subfigure[CIFAR-10.]{
		\label{supp3b}
		\includegraphics[width=0.40\linewidth]{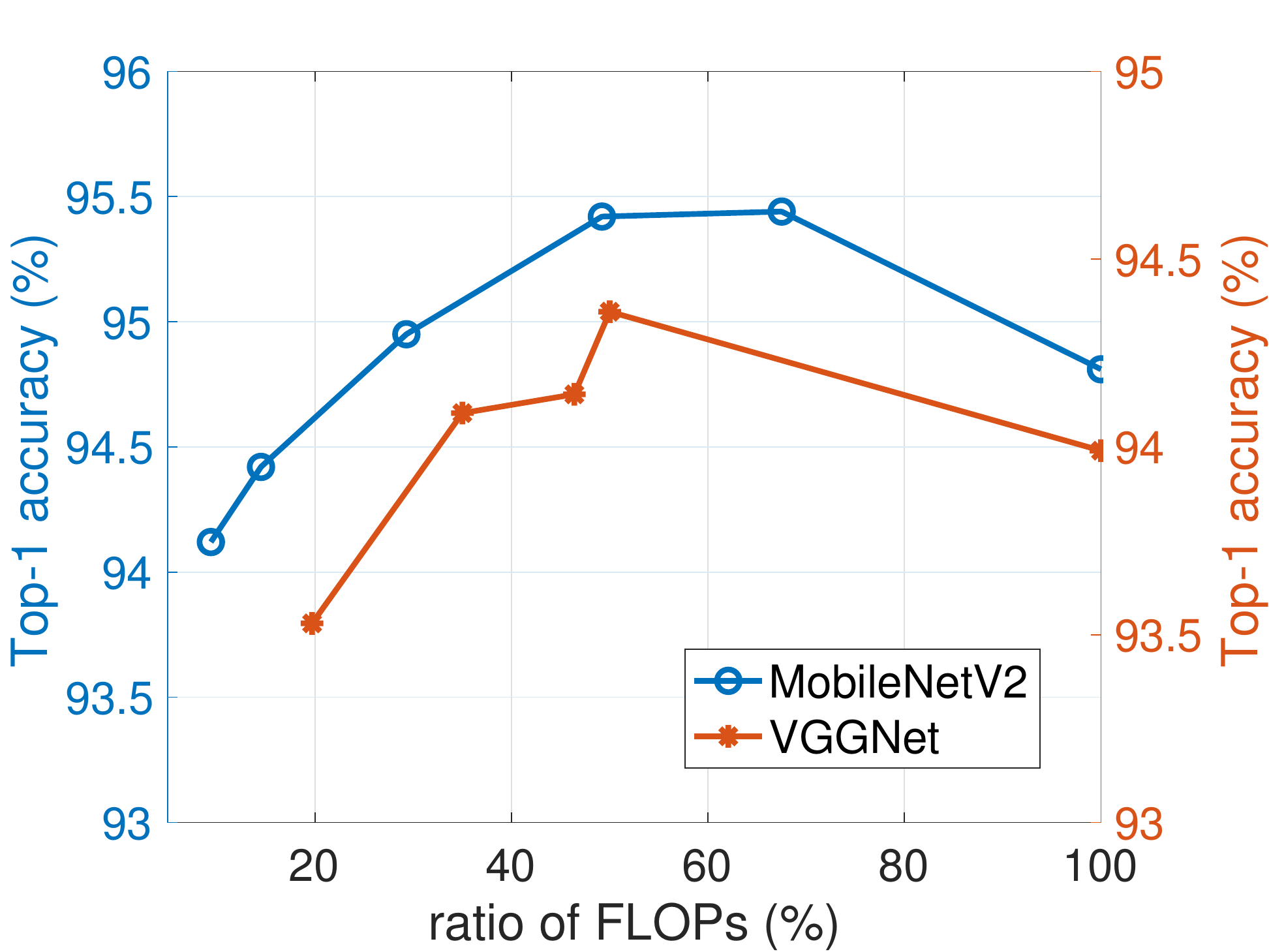}}
	\caption{Top-1 accuracy of searched models on ImageNet and CIFAR-10 dataset \wrt~different ratios of FLOPs.}
	\label{supp3}
\end{figure}

To explore the performance of searched models \wrt~different FLOPs, we include more results on the Top-1 accuracy of ResNet50, MobileNetV2, and VGGNet searched on ImageNet and CIFAR-10 datasets, as shown in Figure \ref{supp3}. Note that we report the ratio of FLOPs of searched models instead of their absolute FLOPs values for clarity. 

Figure \ref{supp3a} shows that the performance of searched models worsens with decreasing ratios of FLOPs on ImageNet dataset. However, the accuracy drop of ResNet50 under a small FLOPs ratio is slighter than that of MobileNetV2. For example, the Top-1 accuracy of ResNet50 (MobileNetV2) drops for 5.2\% (9.9\%) with 14.1\% (16.7\%) ratio of FLOPs. Besides, as shown in Figure \ref{supp3b}, our searched MobileNetV2 and VGGNet can achieve better performance than the original model on CIFAR-10 dataset. In details, our searched MobileNetV2 (VGGNet) with 50\% ratio of FLOPs outperforms the original model (100\% ratio of FLOPs) by 0.63\% (0.37\%) on Top-1 accuracy.

\section{More Detailed Results of ImageNet for Table \ref{Experiments_Imagenet}} \label{A7}
To investigate the effect of BCNet, we further report our algorithm on ResNet34 and ResNet18 with the same training recipe as ResNet50. The original ResNet34 (ResNet18) has 21.8M (11.7M) parameters and 3.6G (1.8G) FLOPs with 74.9\% (71.5\%) Top-1 accuracy, respectively. As shown in Table \ref{Experiments_Imagenet_supp}, our searched 0.5$\times$ ResNet34 (ResNet18) achieve 73.3\% (69.9\%) FLOPs, which exceeds FPGM (DMCP) by 0.8\% (0.7\%) on Top-1 accuracy. Moreover, with tiny FLOPs (\ie, 360M and 450M), BCNet can surpass the unform baseline by a large margin.
\begin{table*}[h]
	\centering
	\scriptsize
	\caption{Performance comparison of ResNet34, ResNet18, and MobileNetV2 on ImageNet. Methods with "*" denotes  that the results are reported with knowledge distillation.}
	\label{Experiments_Imagenet_supp}
	{\begin{tabular}{c|c|cc|cc||c|c|cc|cc}
			\hline
			\multicolumn{6}{c||}{ResNet34} & \multicolumn{6}{c}{MobileNetV2}\\ \hline
			Groups&Methods&FLOPs&Parameters&Top-1&Top-5&Groups&Methods&FLOPs&Parameters&Top-1&Top-5 \\ \hline
			\multirow{7}*{2.7G} & Rethinking	 & 2.79G & - & 72.9\% & -  &\multirow{6}*{150M} & TAS* & 150M & - & 70.9\% & - \\
			& PF & 2.79G & -  & 72.1\% & - && LEGR & 150M & - & 69.4\% & - \\ 
			& MIL & 2.75G & - & 73.0\% & - && Uniform  & 150M & 2.0M & 69.3\% & 88.9\%  \\ 
			& Uniform & 2.7G & - & 72.3\% & 90.8\% && Random & 150M & - & 68.8\% & 88.7\% \\
			& Random & 2.7G & - & 71.4\% & 90.6\% && \textbf{BCNet} & 150M & 2.9M & \textbf{70.2\%}& 89.2\% \\
			& \textbf{BCNet} & 2.7G & 20.2M & \textbf{74.9\%} & 92.4\% && \textbf{BCNet*} & 150M & 2.9M & \textbf{71.2\%}& 89.6\% \\ \cline{7-12}
			& CNN-FCF & 2.7G & 15.9M & 73.6\% & 91.5\% & \multicolumn{6}{c}{ResNet18}\\  \cline{7-12} 
			& \textbf{BCNet} & 2.5G & 20.0M & \textbf{74.6\%} & 92.2\% & Groups&Methods&FLOPs&Parameters&Top-1&Top-5 \\  \cline{1-6} \cline{7-12}
			\multirow{10}*{1.8G} & FPGM & 2.2G & - & 72.5\% & - & \multirow{5}*{1.2G} & TAS*  & 1.2G & - & 69.2\% & 89.2\% \\
			& SFP & 2.2G & - & 71.8\% & 90.3\% && MIL & 1.2G & - & 66.3\% & 86.9\% \\
			& CNN-FCF& 2.2G & 12.6M & 72.8\% & 91.0\% && Uniform  & 1.2G & 8.5M & 68.8\% & 88.5\% \\
			& GS & 2.1G & - & 72.9\% & - && Random & 1.2G & -& 68.4\% & 88.1\% \\
			& Uniform & 1.8G & - & 71.5\% & 90.2\% && \textbf{BCNet} & 1.2G & 11.6M & \textbf{71.3\%}& 90.1\% \\  \cline{7-12}
			& Random & 1.8G & - & 70.9\% & 89.8\% &\multirow{10}*{1G}& SFP & 1.05G & - & 67.1\% & 87.8\% \\
			& \textbf{BCNet} & 1.8G & 16.9M & \textbf{73.3\%} & 91.4\% && FPGM & 1.04G & - & 68.4\% & 88.5\% \\  
			& CGNet & 1.8G & - & 71.3\% & - && DMCP & 1.04G & - & 69.2\% & - \\
			& CNN-FCF & 1.7G & 9.6M & 71.3\% & 90.2\% && CGNet & 0.94G & - & 68.8\% & - \\ \cline{1-6}
			\multirow{5}*{0.9G}& CNN-FCF & 1.2G & 7.1M & 69.7\% & 89.3\% && DCP & 0.96G & - & 67.4\% & 87.6\% \\ 
			& CGNet & 1.2G & - & 70.2\% & - && FBS & 0.9G & - & 68.2\% & 88.2\% \\
			& Uniform & 0.9G & - & 69.6\% & 89.5\% && Uniform & 0.9G & 6.0M & 67.1\% & 87.5\% \\
			& Random & 0.9G & - & 68.8\% & 88.7\% && Random & 0.9G & - & 66.7\% & 87.1\% \\
			& \textbf{BCNet} & 0.9G & 9.6M & \textbf{72.2\%} & 89.8\% && \textbf{BCNet} & 0.9G & 9.8M & \textbf{69.9\%} & 89.1\% \\ \hline
			\multirow{3}*{0.36G}& Uniform & 0.36G & - & 59.6\% & 82.1\% &\multirow{3}*{450M}& Uniform & 450M & 2.9M & 61.6\% & 83.6\% \\
			& Random & 0.36G & - & 56.4\% & 80.7\% && Random & 450M & - & 59.8\% & 82.3\% \\ 
			& \textbf{BCNet} & 0.36G & 3.6M & \textbf{64.4\%} & 85.7\% && \textbf{BCNet} & 450M & 4.9M & \textbf{65.8\%} & 86.4\% \\ \hline
	\end{tabular}}	
\end{table*}


\section{More Visualization of Searched Network Widths} \label{A9}

\begin{figure*}[h]
	\centering
	\includegraphics[width=0.81\linewidth]{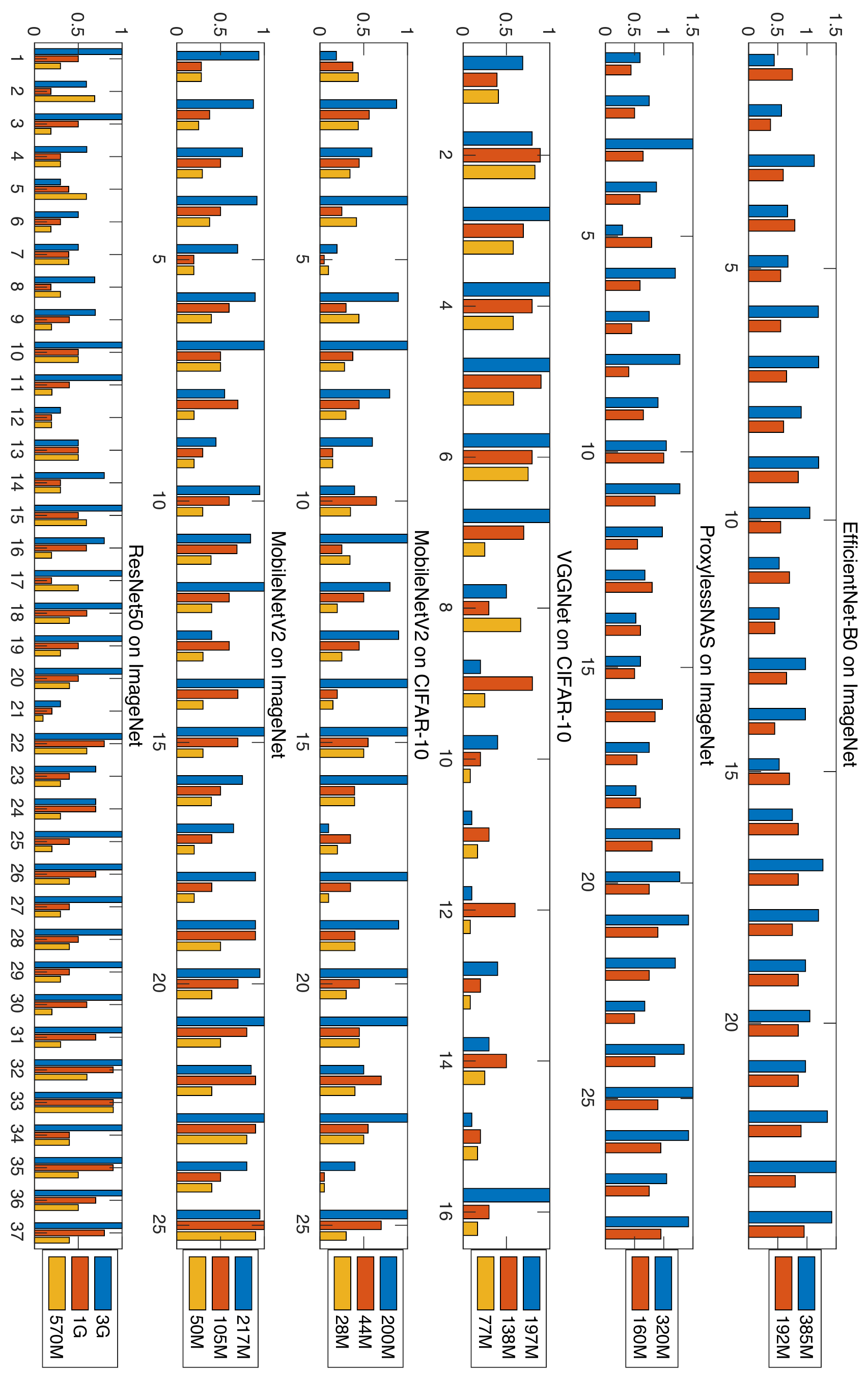}
	\caption{More visualization of searched networks \wrt~different FLOPs. The vertical axis means the ratio of retained width compared to that of original networks at each layer.}
	\label{visualization_supp}
\end{figure*}

For intuitively understanding, we visualize our searched networks as in Figure \ref{visualization_supp}. Moreover, we show the retained ratio of layer widths for clarity compared to that of the original models.  Note that for MobileNetV2, ResNet50,  EfficientNet-B0, and ProxylessNAS with skipping or depthwise layers, we merge these layers, which are required to have the same width. From Figure \ref{visualization_supp}, we notice that the width of EfficientNet-B0 and ProxylessNAS varies more evenly than other models, which means these NAS searched models have more delicate widths than others. Besides, models pruned on CIFAR-10 tend to retain a smaller width on the last layer than on ImageNet, which indicates that the difficulty of the dataset determines the width of the last layer.

\end{document}